\documentclass{article}

\usepackage{PRIMEarxiv}

\usepackage[utf8]{inputenc} 
\usepackage[T1]{fontenc}    
\PassOptionsToPackage{hyphens}{url}
\usepackage{hyperref}       
\usepackage{booktabs}       
\usepackage{amsfonts}       
\usepackage{nicefrac}       
\usepackage{microtype}      
\usepackage{lipsum}
\usepackage{fancyhdr}       
\usepackage{graphicx}       
\graphicspath{{media/}}     

\usepackage{tabularx}
\usepackage{booktabs}
\usepackage{ragged2e}

\usepackage{enumitem}

\usepackage{verbatim}       
\usepackage{xcolor}         
\usepackage{amsmath}        
\usepackage{multirow}      
\usepackage{colortbl}
\usepackage{amssymb}

\newcommand{\yue}[1]{\textcolor{orange}{\textbf{Yue:} #1}}
\newcommand{\harsha}[1]{\textcolor{violet}{\textbf{Harsha:} #1}}

\title{Integrating ESG and AI: A Comprehensive Responsible AI Assessment Framework
}

\author{
  Sung Une Lee, Harsha Perera, Yue Liu, Boming Xia, Qinghua Lu, Liming Zhu \\
  CSIRO's Data61, Australia\\
   \AND
  Jessica Cairns, Moana Nottage \\
   Alphinity Investment Management, Australia \\
}

\begin{document}
\maketitle

\begin{abstract}
Artificial Intelligence (AI) is a widely developed and adopted technology across entire industry sectors. 
Integrating environmental, social, and governance (ESG) considerations with AI investments is crucial for ensuring ethical and sustainable technological advancement. 
Particularly from an investor perspective, this integration not only mitigates risks but also enhances long-term value creation by aligning AI initiatives with broader societal goals.
Yet, this area has been less explored in both academia and industry.
To bridge the gap, we introduce a novel ESG-AI framework, which is developed based on insights from engagements with 28 companies and comprises three key components.
The framework provides a structured approach to this integration, developed in collaboration with industry practitioners.
The ESG-AI framework provides an overview of the environmental and social impacts of AI applications, helping users such as investors assess the materiality of AI use. Moreover, it enables investors to evaluate a company's commitment to responsible AI through structured engagements and thorough assessment of specific risk areas.
We have publicly released the framework and toolkit in April 2024, which has received significant attention and positive feedback from the investment community.
This paper details each component of the framework, demonstrating its applicability in real-world contexts and its potential to guide ethical AI investments.
\end{abstract}


\section{Introduction} \label{sec:introduction}

Environmental, Social and Governance (ESG) represents a framework for evaluating a company's impact on environmental sustainability, social responsibility, and effective governance practices \cite{halid2023literature}.
Investors consider these comprehensive factors to evaluate a company’s impact and performance, rather than focusing solely on financial metrics.

Artificial Intelligence (AI) is a rapidly growing force impacting how companies build new markets, drive productivity improvements and enhance customer engagement.
There has been clearly seen growth in the market, leading to a dramatic increase in company interest. 
Additionally, regulations are evolving, with new local and global AI regulations emerging, such as EU AI Act.
AI presents tremendous opportunities but also poses significant risks, which are typically addressed through Responsible AI (RAI) practices. On the other hand, investors—who may have the greatest leverage to encourage companies to adopt responsible AI—often use the ESG framework for evaluating non-financial metrics. 
There are many overlaps between the concerns and approaches of RAI and ESG. 
The motivation is to integrate responsible AI into the ESG framework, rather than simply managing RAI as a separate entity. 
This integration allows us to view and manage responsible AI through the lens of ESG.

There have been a number of studies on ESG and AI integration, proposing frameworks and protocols for evaluating AI impacts in an ESG context \cite{saetra2021framework, brusseau2023ai}. These studies highlight the need for standardized guidelines, particularly for measuring environmental topics, and emphasize challenges due to diverse standards and complex datasets \cite{Crona2021sweet, yu2024ontology}.
Despite these efforts in academia~\cite{saetra2023ai, minkkinen2024investors} and recent industry reports outlining frameworks for ESG and AI integration~\cite{ESGParadigms, ResponsibleAIPlaybook}, there still lack a comprehensive framework that can be readily implemented in real-world contexts.
We have identified few existing frameworks that specifically support investor decision-making by integrating both ESG and AI considerations. 
More specifically, no frameworks provide tangible and practical toolkits for investors beyond conceptual-level discussions. 
A significant gap exists in the ESG and AI landscape.

Based on collaborative research with dedicated investors and 28 Australian and globally listed companies' engagement, we have developed a new ESG-AI framework which includes three key components such as AI use case, AI governance indicators, and RAI deep dive assessment. We have also provided an enabling toolkit to support the utilization of the framework. 
The framework and toolkit primarily support investors, including ESG managers, but can also be used by AI companies to assess their RAI practices and risks.
Our framework has been designed based on some key existing RAI resources and regulations such as the RAI question bank \cite{lee2023qb4aira}, RAI metric catalogue \cite{xia2023principles}, Australia AI Ethics Principle\footnote{https://www.industry.gov.au/publications/australias-artificial-intelligence-ethics-framework/australias-ai-ethics-principles\label{AUprinciple}}, EU AI Act\footnote{\url{https://artificialintelligenceact.eu}\label{euaiact}}, NIST AI Risk Management Framework\footnote{\url{https://www.nist.gov/itl/ai-risk-management-framework}\label{nist}}, AI Standard (ISO/IEC 42001)\footnote{\url{https://www.iso.org/standard/81230.html}\label{iso42001}} and more.
This draws on insights and standards from key regulatory bodies, standard organisations and stakeholder groups.

There are three key contributions of this study. Firstly, it delivers comprehensive insights on the integration of ESG and AI, particularly from the investor perspective, a topic that has been explored to a limited extent previously. Secondly, it provides practical tools for practitioners by operationalizing AI Ethics principles, bringing them from an abstract to an implementation level. 
Through iterative testing and evaluation, these tools have been designed for adoption by the investment community and companies as standard practice for RAI measurement.
Lastly, our framework, along with a set of RAI metrics, encourages companies to measure, manage, and disclose their performance data transparently. Metrics for RAI and ESG are not often shared publicly, making it challenging for investors to assess performance and understand what best practices look like.

The remainder of this paper is as follows. 
The next section provides background information of the previous studies on ESG and AI studies, relevant RAI frameworks and regulations used in this study and key ESG topics associated with the AI ethics principles.
Section~\ref{sec:methodology} describes our approach for design and development the framework.
We then introduce the ESG-AI framework in Section \ref{sec:framework} including overall architecture and the key components of the framework in detail.
Section \ref{sec:discussion} presents how our framework has gained users' attention since its release, and discuss the practical implications of this study. 
We conclude this study in Section~\ref{sec:conclusion}.


\section{Background and Literature Review} \label{sec:background}




In recent years, ESG frameworks have emerged as an overarching framework to assess enterprises across three dimensions:
Environmental (e.g., carbon emissions, water usage), Social (e.g., employee diversity, working conditions), and Governance (e.g., board composition, audit practices).
ESG topics can be connected to AI in three ways: leveraging AI to reduce ESG risks (e.g., emission reduction, monitoring, and governance); using AI for positive impacts (e.g., customer experience); and addressing the concerns and risks of RAI, which overlap with ESG concerns and risks.
Integrating AI into ESG frameworks is crucial for addressing these risks and concerns.

While widely adopted ESG frameworks, such as the Global Reporting Initiative standards\footnote{https://www.globalreporting.org/standards/\label{esgstandards}}, provide high-level guidance, Crona \cite{Crona2021sweet} and Yu et al. \cite{yu2024ontology} highlighted the lack of standardized guidelines for measuring ESG metrics, especially environmental aspects. What is more, AI, integral to modern enterprise workflows, enhances ESG analysis and operationalization. TSE et al. \cite{tse2023ai} describe AI-enabled ESG tools involving three main steps: harvesting, which involves collecting and parsing ESG-related data; organizing, which entails screening and transforming data into structured formats; and analyzing, which includes performing classification, sentiment, contextual, and semantic analysis.
Pozzi and Dwivedi~\cite{pozzi2023esg} analyse the integration of IoT and ESG, especially environmental sustainability, and they propose a reference architecture where devices' real-time data are collected and monitored, and an alarm will be triggered if a predefined threshold is met. 
The authors also discussed the positive impacts of using AI on ESG, including efficient and accurate data collection, analysis, and decision-making.
Xu~\cite{xu2024ai} conducted a survey to investigate how AI systems facilitate the analytical capabilities, risk assessment, and customer engagement of ESG in financial institutes. 

While AI can enhance ESG, it also requires additional considerations within the ESG framework to address its own associated risks and concerns.
For example, researchers have studied AI's impact on sustainability~\cite{vinuesa2020role, tomlinson2024carbon}. Saetra~\cite{saetra2021framework} proposed a framework for evaluating AI impacts using the United Nations’ Sustainable Development Goals. More recently, Saetra~\cite{saetra2023ai} introduced an AI ESG protocol for evaluating and disclosing AI's ESG impacts, comprising four main steps: initial descriptive statement, main impact statement, risks and opportunities, and action plan. Brusseau \cite{brusseau2023ai} suggested nine performance indicators to score AI human impact, which mainly focuses on social aspects such as personal freedom, social wellbeing and technical trustworthy. Crona~\cite{Crona2021sweet} and Yu et al.~\cite{yu2024ontology} highlighted the lack of standardized guidelines for measuring ESG metrics, especially environmental aspects. Minkkinen et al.~\cite{minkkinen2024investors} explored AI auditing through ESG dimensions via interviews, focusing on AI awareness, impact measurement, and governance. Khoruzhy et al.~\cite{khoruzhy2022esg} compared the ESG investments in AI regarding a set of developed and developing countries.

With advancements in AI (e.g., Large Language Models), there are concerns about whether such (advanced) AI systems can behave as intended, in a responsible and safe manner, which results in the emergence of \textit{responsible AI}~\cite{RAIbook} and AI Safety\footnote{\url{https://www.gov.uk/government/publications/ai-safety-summit-2023-the-bletchley-declaration/the-bletchley-declaration-by-countries-attending-the-ai-safety-summit-1-2-november-2023}}. RAI is inherently aligned with ESG as it involves both legal and broader ethical considerations. The Responsible AI Institute devised a framework integrating RAI into ESG paradigms~\cite{ESGParadigms}, and a responsible AI playbook for investors was published in the World Economic Forum 2024~\cite{ResponsibleAIPlaybook}.

Governments and industry have also attached great importance to ESG-AI and RAI in general. The EU Council approved EU AI Act\textsuperscript{\ref{euaiact}} in May 2024, defining four levels of risks (unacceptable, high, limited, and minimal) and specifying legal requirements for general-purpose AI. Similarly, the US National Institute of Standards and Technology developed an AI risk management framework\textsuperscript{\ref{nist}} for identifying, assessing, and mitigating AI-related risks. The ISO/IEC 42001 Standard\textsuperscript{\ref{iso42001}} provides guidelines for enterprise leadership and governance boards on establishing and maintaining AI management systems.
Moreover, frameworks such as the OECD Framework \footnote{{\url{https://www.oecd.org/en/publications/oecd-framework-for-the-classification-of-ai-systems\_cb6d9eca-en.html}\label{oecd}}}for the Classification of AI Systems and the Microsoft Responsible AI Standard~\footnote{\url{https://blogs.microsoft.com/wp-content/uploads/prod/sites/5/2022/06/Microsoft-Responsible-AI-Standard-v2-General-Requirements-3.pdf}\label{msstandard}} incorporate ESG dimensions, often adopting the "Triple Bottom Line" approach (people, planet, profit). In general, all these RAI frameworks embody ESG consideration.
For instance, the Australian government’s AI Ethics Framework\textsuperscript{\ref{AUprinciple}} includes principles like `\textit{Human, social and environmental wellbeing}'', directly relating to ESG topics such as greenhouse gas emissions and resource efficiency.
RAI is considered within ESG topics as follows (detailed mapping between ESG topics and RAI principles is presented in Table \ref{tab:ESG mapping}): 





\textbf{Environmental:}

\begin{itemize}
    \item \textit{Greenhouse gas (GHG) emissions}: AI model training, deployment and operation may require a significant amount of energy, whilst, AI systems can help reduce GHG emissions via asset optimisation, workflow automation and operational efficiency.

    \item \textit{Resource efficiency}: AI can optimise resource efficiency across the supply chain, hence adopting AI systems can help reduce energy, land and water consumption. 

    \item \textit{Ecosystem impact}: AI systems can leverage big data to monitor and address key environment ecosystem challenges such as deforestation, soil health and pollution.
\end{itemize}

\textbf{Social:}

\begin{itemize}
    \item \textit{Diversity, equity, and inclusion}: On one hand, AI models may suffer or even introduce new forms of discrimination during training. On the other hand, using up-to-date, high‑quality, and diverse datasets for training can ensure the support of diversity and inclusion.

    \item \textit{Human rights}: Particular use of AI (e.g. surveillance, weapons, misinformation dissemination) can breach human rights. Instead, AI systems can improve supply chain transparency and information sharing, and the use of robotics to accomplish low-value and unsafe tasks help protect human rights.

    \item \textit{Labour management}: Enterprises can adopt AI to automate repetitive tasks and improve productivity outcomes, hence addressing the issue of labour shortages. Nevertheless, this may result in job losses, especially for lower-paid roles.

    \item \textit{Customer and community}: A company may suffer reputational risks in AI safety, accountability, reliability and explainability, while it can also benefit from enhanced product quality, better customer service, and recognised leadership related to AI opportunities.

    \item \textit{Data privacy and cybersecurity}: Leveraging big data for AI model training requires consideration on data privacy, consent, fraud and security. Meanwhile, AI systems can facilitate cybersecurity via fraud detection and predictive analysis.

    \item \textit{Health and safety}: AI systems can support hazard prediction and recognition to prevent and minimise high-severity injuries, which is dependent on system operation and automation.
\end{itemize}

\textbf{Governance:}

\begin{itemize}
    \item \textit{Board and management}: Leadership awareness and commitment are significant for successful AI adoption.

    \item \textit{Policy (internal and external)}: Responsible AI policies can be referred to as an indicator for AI leadership and operationalise ethical AI practices.

    \item \textit{Disclosure and reporting}: ESG disclosure can help promote RAI disclosure, which are both significant for companies to maintain a strong social licence and ensure transparency with stakeholders.
\end{itemize}


Nevertheless, we observed that most existing works on ESG-AI focus primarily on environmental topics and often neglect the investor's perspective, which is crucial for applying ESG frameworks. 
In particular, Saetra~\cite{saetra2023ai} proposes an protocol for enterprise directors and managers to evaluate the ESG impacts regarding AI and data capabilities, assets, and activities. Minkkinen et al.~\cite{minkkinen2024investors} report their interview results with Finnish senior-level experts about the awareness of AI issues, AI impact measurement and governance in the context of ESG. The guiding framework on integrating AI into ESG~\cite{ESGParadigms} includes only three metrics for each ESG dimension: greenhouse gas emissions, energy consumption, and water consumption for environmental metrics; health and safety, diversity, equity, and inclusion, and human rights and labor standards for social metrics; and board composition and oversight, compliance, and Codes of Conduct, and governance and accountability for governance metrics. However, this framework lacks further analysis and detailed rubrics.
Similarly, the RAI playbook for investors~\cite{ResponsibleAIPlaybook} analyses AI risks and stakeholders' roles but focuses more on the engagement process with companies, without providing tangible tools for investors to work with.
Consequently, there is an urgent need for a holistic framework that integrates both ESG and AI considerations for investors across various real-world contexts. This paper proposes a comprehensive framework, consisting of a set of AI use cases, 10 RAI governance indicators, and in-depth assessments for operationalizing high-level principles and guidelines. The proposed solution can serve as an assessment tool for the investment community.

We compared this study with the related work based on ESG topics, AI ethics principles, target users, key elements, and toolkit support. 
For ESG topics, we used the 12 ESG topics described above; the comparison of AI ethics principles is based on Australia eight AI ethics principles (Human, societal, environmental wellbing, Human-centred values, Fairness, Privacy and Security, Reliability and safety, Transparency and explainability, Contestability and Accountability). 
The results are demonstrated in Table~\ref{tab:relatedwork}. 
This comparison highlights that this study comprehensively covers the integration of ESG considerations and AI ethics principles, providing a robust toolkit for investors.

\begin{table*}[!tbhp]
\footnotesize
\centering
\caption{Comparison with the related work in academia and industry.}
\label{tab:relatedwork}
\begin{tabular}{p{0.1\textwidth}p{0.1\textwidth}p{0.05\textwidth}p{0.2\textwidth}p{0.1\textwidth}p{0.2\textwidth}p{0.05\textwidth}}
\hline
Related work & ESG topic & \multicolumn{2}{l}{AI principle (exception)} & Target & Focus & Toolkit \\
\hline

\cite{saetra2023ai} & Fully & Partially & (Tranparency/explainability, Fairness, Contestability) & Company & AI and data driven impacts- the AI ESG protocol structure. & No \\

\cite{minkkinen2024investors} & Fully & Partially & (Human-centred values, Reliability/safety, Fairness, Contestability) & Investor & Awareness of responsible use of AI and the connection with ESG. & No \\
\cite{ESGParadigms} & Fully & Partially & (Tranparency/explainability, Priavcy/security, Contestability) & Company & ESG metrics and a framework for leveraging AI to advance achieving ESG Goals. & No \\

\cite{ResponsibleAIPlaybook} & Fully & Partially & (Contestability, Accountability) & Investor & AI risks and role of stakeholders, and investor engagement in RAI and hurdles. & No \\

\textbf{This study} & \textbf{Fully} & \textbf{Fully} & \textbf{-} & \textbf{Investor} & \textbf{A comprehensive ESG-AI framework and investor assessment toolkits.} & \textbf{Yes} \\
\hline

\end{tabular}
\end{table*}

\section {Methodology} \label{sec:methodology}

\begin{figure*} [htb]
    \centering
    \includegraphics[width=\textwidth]{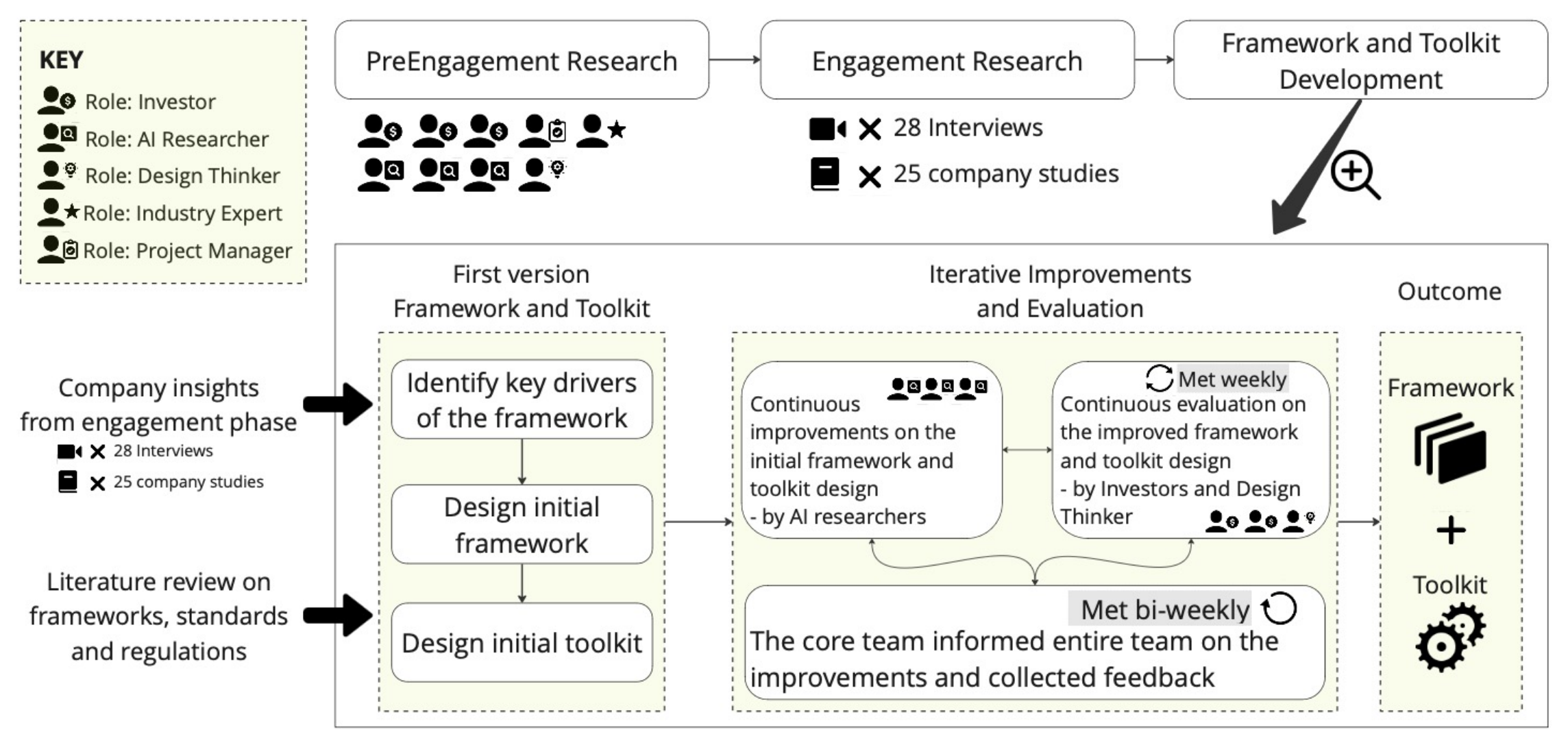}
    \caption{The methodology - focusing on the framework and toolkit development}
    \label{fig:methodology}
\end{figure*}

We adopted a collaborative research methodology~\cite{ward1982collaborative} to develop the framework and toolkit, focusing on continuous dialogue and goal alignment. 
In this study, we considered the three key factors of a collaborative research.
First, researchers and practitioners work together. Second, we focus on "real word" as well as theoretical problem. Third, the participants gain mutual respect for one another and grow in their insights into and understanding of ESG and AI.
This research was implemented through three distinct phases: Pre-engagement Research, Engagement Research, and Framework \& Toolkit Development (Figure \ref{fig:methodology}). 
This paper focuses on the Framework \& Toolkit Development phase, as the framework and toolkit are the key deliverables of the research. While we will briefly outline the methodologies used in the first two phases, the focus of this section will be the third phase, which provides detailed context for the framework development.

\subsection{Pre-engagement Research Phase (Feb 2023 - Apr 2023)}
\textbf{\textit{Formation of the Research Team:}}


The research team was assembled, comprising three AI researchers, three ESG experts/investors, a senior design thinker, a project manager, and a senior industrial expert. Each member had a specific role (Table \ref{tab:researchteam}).

\begin{table} [htb]
 \footnotesize
    \begin{tabular}{p{0.2\textwidth}p{0.75\textwidth}}
    \hline
   Role& Description\\ 
    \hline
         AI Researcher& Provided insights into AI principles, technologies, and responsible AI practices. Drive the framework and toolkit development\\
         ESG Expert/Investor & Shared perspectives on investment strategies, ESG considerations and assessment needs.\\
         Design Thinker & Ensured the framework's usability from a user-centered design perspective.\\
         Project Manager & Coordinated the project activities and kept the research team on schedule.\\
         Senior Industrial Expert & Contributed industry-specific knowledge and practical insights while balancing out different perspective from AI researchers and investors \\
    \hline
    \end{tabular}

    \caption{Research team - roles and responsibilities}
    \label{tab:researchteam}
\end{table}

Initial meetings were held to establish common goals, objectives, and research questions. These discussions facilitated a shared understanding of responsible AI principles and assessment needs of investors, specifically related to ESG considerations. Further, the team conducted a comprehensive review of existing literature on responsible AI and investment frameworks to identify the best practices (e.g.,~\cite{saetra2021framework, brusseau2023ai}), standards~\textsuperscript{\ref{iso42001}, \ref{esgstandards},\ref{msstandard}}, and regulatory requirements~\textsuperscript{\ref{euaiact}}.

\textbf{\textit{Interview Protocol and Questionnaire:}}
The team conducted a workshop to identify the target sectors and companies for interviews on RAI practices and their intersection with ESG criteria. Investors and the senior industry expert played a critical role in prioritizing industries and selecting companies for interviews.
An initial questionnaire was developed and subsequently revised. Investors were responsible for formulating questions related to ESG aspects, while AI researchers focused on developing questions relevant to RAI. The initial questionnaire was then tailored slightly for each interviewee-company to ensure the questions were aligned with the specific sector and the product or service they represented. For example, companies from the Information Technology (IT) sector, known for their leadership in developing RAI frameworks, were posed questions tailored to their advanced understanding of RAI.

\subsection{Engagement Research Phase (Apr 2023 - Sep 2023)}
\textbf{\textit{Interviews:}} 
We engaged directly with 28 companies across 8 sectors, including leading global and Australian companies. Notably, 34\% of the invited companies declined participation in the interviews. Reasons cited for declining included their low level of AI maturity or concerns about market sensitivity. As per the interview protocol, one investor conducted the interviews, focusing on broader aspects, while an AI researcher handled RAI-related questions. Other team members participated as observers and note takers. The interviews were not recorded due to the privacy and sensitive information collected.

\textbf{\textit{Data Analysis and Insights:}} 
Once she interviews were completed, the collected data and researcher notes were shared among the team members. The shared data was analysed separately by the investors and AI researchers to identify the best practices, the level of RAI maturity of the companies and the key insights from all the companies we interviewed. Then, the team conducted a series of workshops to synthesis the outcomes of investors and AI researchers to come up with a collective output. 
Following the completion of interviews, the gathered data and researcher notes were distributed among team members. Investors and AI researchers independently analyzed the shared data to discover best RAI practices, assess the RAI maturity levels of companies, identify the intersection between RAI \& ESG and extract key insights from all interviewed companies. Subsequently, the team conducted a series of workshops to synthesize the findings of both investors and AI researchers into a unified output.
The final insights were used as one of the key inputs of the framework development phase; see Section~\ref{subsec:Company insights into the framework design}. 

\subsection{Framework Development Phase (Sep 2023 - Mar 2024)}
\subsubsection{Framework Design}
When designing the framework, we mainly input the insights from the previous phase of the project and the knowledge gathered using literature review. We explain how this input influence the design of the framework in details in Section~\ref{sec:framework}. Using these inputs we identified the key drivers of the framework. 
\begin{itemize}
    \item \textit{\textbf{Industry vs. Company:}} 
    We observed significant similarities in Responsible AI (RAI) practices and use cases among companies within the same industry (e.g., financial sector). Therefore, the impact RAI on investment decisions may initially be analysed at the industry level and subsequently drilled down to the company level.
    
    \item \textit{\textbf{The level of accessibility to information of a company:}}  
    Investors and portfolio managers can readily access publicly available information about both industries and individual companies. Moreover, we found that investors show equal interest in governance indicators specific to companies. This information may gather engaging directly with the company through methods such as interviews.
    
    \item \textit{\textbf{The intersection of ESG and RAI principles:}} ESG and RAI principles exhibit overlapping characteristics, with key indicators and metrics aligning in certain areas. This alignment serves as a crucial bridge between investor knowledge and RAI principles. We have discussed this in Section \ref{sec:discussion}; see Table \ref{tab:ESG mapping}.
    \item \textit{\textbf{A questionnaire:}} Further, a framework that supports investors to assess the RAI of a company to make investment decisions should thoroughly examine company practices. Such evaluation requires a carefully constructed questionnaire that aligns with current frameworks, standards, and regulatory guidelines.
\end{itemize}
\subsubsection{Framework Development, Iterative Improvements and Evaluation}
Using the key elements mentioned above, AI researchers developed the initial version of the ESG-AI framework and toolkit. The entire team was then informed about the framework, and their feedback was collected. A core team, consisting of selected investors, AI researchers, and a design thinker, was subsequently formed to further refine the framework. 
AI researchers in the core team was responsible on continuously improving the framework and the toolkit. The investors evaluated the improved versions of the framework and toolkit with their workflow and informed the core team on any required adjustments and further suggestions. Unlike other similar frameworks~\cite{ResponsibleAIPlaybook}, which is tested after the development, this framework was evaluated with real-world clients, the investors in this case, continuously and collaboratively.  

The core team met weekly to discuss potential improvements and reported updates to the broader research team bi-weekly, formally documenting the feedback received from the rest of the team. This process continued for nearly five months, underscoring the thoroughness of the iterative revisions made to the framework.

\textit{The role of investors' collaboration:} Aligning with the collaborative research guidelines~\cite{ward1982collaborative}, the investors in the research team played a crucial role, as they represented the end users of the framework. Many design decisions were proposed or adjusted to accommodate the requirements or workflows suggested by the investors. For example, when designing the framework's assessment module, investors recommended incorporating specific governance indicators that they frequently use in their evaluations, particularly in ESG. This input ensured that the framework not only aligned with technical standards but also integrated practical considerations from the investor's perspective.

After multiple rounds of feedback and thorough validation from investors and the entire team, the framework and toolkit were finalized. 

\subsubsection{Publication and Outreach}
The final stage of our collaborative effort involved sharing the completed framework and toolkit through medias, industry reports and online platforms. To promote the framework, we organized conference presentations, webinars, and engagements with industry stakeholders. 
Further, a dedicated sanity URL was introduced with a web page \footnote{\url{https://www.csiro.au/en/research/technology-space/ai/responsible-ai/rai-esg-framework-for-investors}}. 
The report is freely available and the toolkit has restricted access to gather the information of people who would like to share their contact information download the toolkit templates. These activities aimed to raise awareness and encourage the adoption of our framework and allow us to contact them for future research. 



\section{ESG-AI framework} \label{sec:framework}

This section introduces the ESG-AI framework we have developed in partnership with the dedicated investors.
First, we present the key insights from our engagement with 28 companies and explain how these insights informed the development of the framework.
We then provide an overview of the framework architecture and its three key components: AI Use Case, RAI Governance Indicators, and RAI Deep Dive Assessment.

\subsection{Company insights into the framework design}
\label{subsec:Company insights into the framework design}
We have identified six key RAI insights for our framework design from the interviews with the engaged companies.

\textbf{\textit{IN1. Employee engagement is essential to deliver AI‑related opportunities.}}
Successful AI implementation requires input from both technical and non-technical staff. For example, engineers and consultants need to generate AI‑related ideas for developers so the business needs can be met effectively. These types of partnerships are particularly crucial in industries like industrial and mining, where technology adoption has traditionally been limited.

Recognizing the significance of employee engagement from diverse team in the successful deployment of AI \cite{smith2019designing}, we have integrated this concept into our framework design. Specifically, we have included "diversity and inclusion" as a key social impact factor. This inclusion ensures that the diverse perspectives and expertise of employees across various roles are considered in AI development and implementation processes \cite{hughes2019artificial}. 

\textbf{\textit{IN2. Strengthening Board and leadership capability in AI, technology and ethics.}}
Directors need tech know-how to navigate AI. Given the competitive landscape for experienced AI directors, alternative approaches such as training and raising awareness among existing Board members become essential but are yet to be fully explored. Companies with technology expertise are better placed to expand knowledge appropriately in the AI space.

The role and competence of boards in successful implementation of RAI has been emphasized by researchers as well \cite{torre2019ai}.
Moreover, in some AI regulations and standards such as ISO/IEC AI management, top management is required to demonstrate leadership and commitment with respect to the AI management. 
The EU AI Act also includes an accountability framework as one of the key requirements for high-risk AI. This encompasses setting up the responsibilities of the management and other staff including boards with regard to all aspects of AI.
We took this concept as a top priority and included "Board and management" topic in the 12 standard ESG topics. 
Breaking down the concept, our framework accommodates it and includes 
"board accountability and board capability" indicators under "board oversight" in RAI governance assessment. 

\textbf{\textit{IN3. RAI governance is best embedded within existing systems and processes.}}
This involves implementing governance structures that involve representatives from various disciplines to analyse risks and make informed decisions about AI strategy aligned with business objectives.
In addition, there is need for defined RAI responsibility and sensitive use cases. For example, \textbf{Microsoft} is recognised for its robust RAI governance structure and leading RAI framework, particularly in explicitly referencing sensitive use cases.

This insight has been mainly used in designing AI governance indicators, especially, for "Sensitive use cases" under "RAI commitment" category and "Dedicated RAI responsibility" and "System integration" of "RAI implementation" category.

\textbf{\textit{IN4. A balanced view of threats and opportunities is needed to mitigate harm and leverage AI benefits.}}
Many companies express concerns about the potential negative impacts on their reputation, consumer trust, and regulatory consequences. While caution is understandable, this should not stifle innovation and the potential for productivity gains. For example, some companies restricted employees from using AI tools such as ChatGPT, while others took an educational stance.

A large number of studies have addressed the opportunity and/or challenge of AI \cite{kasneci2023chatgpt}. 
Since \textit{Generative AI} emerging (e.g., ChatGPT), this has been accelerated in both industry and academia.
Most RAI frameworks underscore the importance of both positive and negative impact assessment, on human, social and environment. 
We also considered this insight in designing our framework to support investor's decision-making.
Particularly, our framework enables investors to analyze environmental and social impact from both opportunities and threats aspects and determine the materiality level of the AI applications across nine industry sectors.

\textbf{\textit{IN5. Companies are using different strategies for navigating and managing RAI risk, but supply chain management can be overlooked.}}
Many of the companies interviewed had not considered managing risks through procurement. Addressing this gap is crucial, especially for sectors that are less tech savvy, to establish an ethical AI ecosystem that goes beyond individual organisational boundaries.
40\% of interviewed companies had internal RAI policies. However, most of the companies have not included the entire AI supply chain in their policies yet.

We regarded "supply chain management" for RAI as a key risk management topic and included management of third-party risks or other supply-chain issues in the RAI deep dive assessment of the framework.

\textbf{\textit{IN6. Data privacy is a key ESG issue, but other topics are still important and may be overlooked.}}
We have identified that data privacy is the most common concern of companies. During the interviews, data privacy and cyber-security were consistently identified as the issues most material to AI.
Meanwhile, human rights and modern slavery were not identified as concerns in the interviews. With one of the core AI ethics principles focusing on human rights, this topic remains critical, yet underexplored in the AI space.

To address this issue, we have adopted the Australia AI ethics principle as overarching category for the framework, which broadly covers RAI topics. 
By doing so, investors can have comprehensive assessment breath, not only privacy concerns but also others such as human rights, accountability, transparency, fairness and more.

Table \ref{tab:insights} shows how our framework has been designed based on the six RAI insights.

\begin{table}[htb]
    \centering
    \footnotesize
    \begin{tabular}{p{0.35\textwidth}p{0.6\textwidth}}
    \hline
    RAI insight & The design concept and principle of framework \\ 
    \hline
    IN1. Employee engagement is essential to deliver AI-related opportunities & Select \textbf{Diversity and inclusion} as a key social impact factor,  include \textbf{employee awareness} for company's commitment assessment and consider \textbf{diverse team involvement} as a key RAI practice. \\
    IN2. Strengthening Board and leadership capability in AI, technology and ethics & Include \textbf{board and management} in the standard ESG topic and use \textbf{board accountability and board capability} indicator to assess "board oversight" of the company. \\
    IN3. RAI governance is best embedded within existing systems and processes & Acknowledge that \textbf{sensitive use cases} is a key indicator for "RAI commitment" and include \textbf{dedicated RAI responsibility} and \textbf{system integration} for "RAI implementation" assessment.\\
    IN4. A balanced view of threats and opportunities is needed to mitigate harm and leverage AI benefits & Enable AI use case analysis from different perspectives- both \textbf{positive and negative impacts on ESG}.\\
    N5. Companies are using different strategies for navigating and managing RAI risk, but supply chain management can be overlooked & Consider \textbf{supply chain issues as key aspects of cyber-security and risk management}, and assess how companies address these concerns in their RAI practices. \\
    IN6. Data privacy is a key ESG issue, but other topics are still important and may be overlooked & Adopt \textbf{comprehensive AI ethics principles}, including privacy, in the framework, and thoroughly assess the RAI practices of companies. \\
    \hline
    \end{tabular}
    \caption{Company insights as key inputs for the framework.}
    \label{tab:insights}
    \normalfont
\end{table}

\subsection{ESG-AI framework: Overview of the structure}

This comprehensive framework comprises three essential components that collectively form the backbone of our ESG-AI framework. These components include an in-depth review of AI use cases for industry sectors, an evaluation of RAI governance indicators, and RAI deep dive assessment for a meticulous examination of RAI principles. Together, these elements provide a holistic approach to assessing and mitigating the risks associated with AI deployment, fostering responsible AI practices across diverse sectors.

\begin{figure*} [htb]
    \centering
    \includegraphics[width=\textwidth]{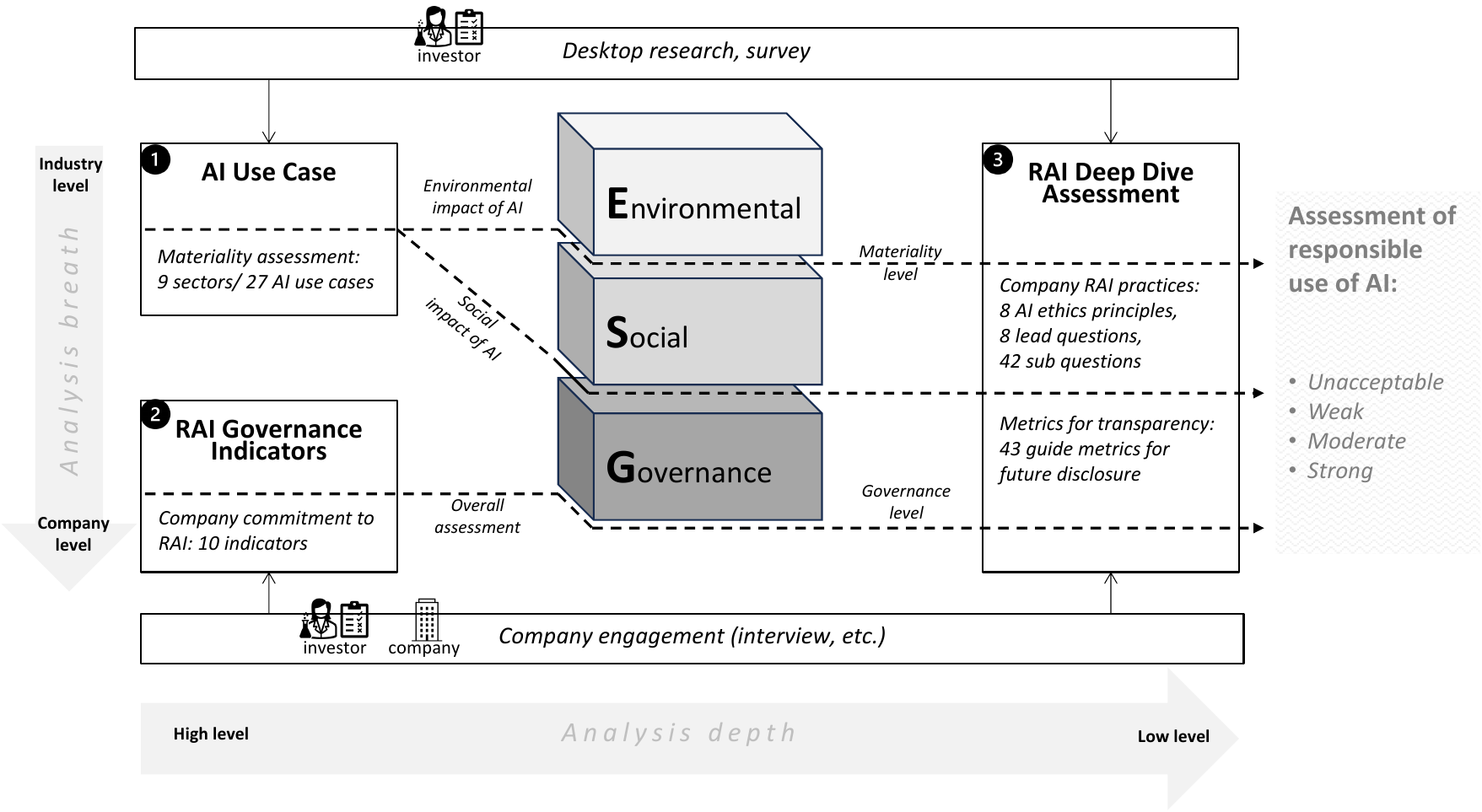}
    \caption{The overview of the ESG-AI framework.}
    \label{fig:framework}
\end{figure*}

Figure \ref{fig:framework} illustrates the overview of the framework. It also demonstrates how the components of the framework are used for assessment of responsible use of AI.

Investors may begin with AI Use Case to gain insights and understand key AI applications across industry sectors. 
This provides an overview of the environmental and social impacts of AI applications in real-world contexts.
Ultimately, investors can determine the materiality level of the AI applications, which may be useful for preliminary sensing potential risks and opportunities associated with the use of AI.

RAI Governance indicators can be used to assess company's commitment to RAI.
This can be implemented through company engagements such as interviews and surveys. 
It provides a high-level governance score, which can be used to identify specific risk areas that need thorough assessment.

The final step, the RAI Deep Dive Assessment, involves a detailed examination of a company's RAI practices. 
This assessment consists of 8 AI ethics principles and 42 questions, along with 42 guiding metrics. 
The final score is categorized into four levels: \textit{Unacceptable, Weak, Moderate, and Strong}.

The next sections describe each components in detail.

\subsection{ESG-AI framework: AI Use Case}

In the Industry 4.0 era, AI is regarded as an essential technology for maximizing performance, product quality and employee well-being \cite{javaid2022artificial}.
The trend shift leads that AI applications have been used everywhere in our daily life, in every industry sectors.
To do preliminary screen and identify companies exposed to material AI use cases, investors need to understand the AI use cases in the industry sectors which are in their investment portfolios.

With such motivation and investor needs, this component, \textit{AI use case} is designed to identify high material AI use cases across industry sectors, from the investor perspective. 
We defined nine industry sectors which are based on the standard industry sectors of the Australian share market \footnote{https://www.listcorp.com/asx/sectors/}, after some modifications.
We have identified the top three AI applications for each sector from a survey of investor AI reports and media coverage, followed by delving into the AI applications.

Table \ref{tab:sector-use case} presents the nine sectors and the selected AI use cases used in this study. Yet, industry sectors and AI use cases are not limited to this. Investors can flexibly add and/or modify any sectors/AI use cases to analyze for their current or future investment portfolios.

\begin{table}[htb]
    \centering
    \footnotesize
    \begin{tabular}{p{0.2\textwidth}p{0.2\textwidth}p{0.5\textwidth}}
    \hline
    Sector & AI use case & Description \\ 
    \hline
    Information technology & Product development & AI tools for new products or using AI to power existing features.\\
     & Automation & Automation of tasks using AI in business processes. \\
     & Risk management & AI to predict when a system might fail or identify vulnerabilities. \\
    Health care & Health research / testing & AI aids in the generation of valuable insights and expediting various processes. \\
     & Clinical care & AI for synthesising and summaries of patient records, early diagnosis or identification of test results. \\
     & Product development & AI for the innovation of health applications and medical devices.\\
    Financials & Insurance pricing & AI can model and calculate prices and terms for insurance products. \\
     & Fraud detection & AI can detect and prevent fraud, protecting customers and banks.\\
     & Credit scoring / approval & AI can augment decisions around who gets access to capital and how much it costs them. \\
    Consumer Discretionary & supply chain management & AI can optimise logistics, predict demand, and improves quality control. \\
     & Personalised offering & AI personalises shopping experiences and advice. \\
     & Instore surveillance & AI can be used to analyse and derive insights from in-store. \\
    Industrials & Process automation & AI is significantly advancing process improvement and automation. \\
     & Asset maintenance & AI can enable real-time equipment monitoring, predicting failures and more. \\
     & Logistics management & AI is optimising delivery routes, automating warehouse operations, and providing precise demand forecasting. \\
    Energy & Energy efficiency & AI-driven systems in homes or businesses equipped with IoT devices and smart meters can manage energy consumption effectively. \\
     & Infrastructure maintenance & AI predicts when equipment in power plants or on the grid might fail and recommending pre-emptive maintenance. \\
     & Energy optimisation & AI can predict energy demand and supply fluctuations. \\
    Real Estate & Property valuation & AI algorithms can estimate property values more accurately and efficiently. \\
     & Facility management & AI algorithms work together to manage building operations such as heating and ventilation. \\
     & Customer services & AI can transform customer by chatbots and virtual assistants. \\
    Materials & Material discovery & AI accelerates the discovery and development of new materials by analysing complex chemical and physical data. \\
     & Resource identification & AI is analysing geological data, satellite imagery, and sensor data from exploration sites to identify promising areas for resource extraction more accurately and quickly. \\
     & Health and safety & AI can use sensing devices to detect unsafe practices or environmental conditions. \\
    Telecommunications & Asset management & AI can predict network traffic and optimise the flow of data. \\
     & Customer service & AI can provide 24/7 customer support, handle routine inquiries, and troubleshoot common issues. \\
     & Fraud detection & AI algorithms can analyse vast amounts of call and data transfer records in real-time. \\
    \hline
    \end{tabular}
    \caption{Selected sectors and the AI use cases for materiality analysis.}
    \label{tab:sector-use case}
    \normalfont
\end{table}

\begin{figure*} [htb]
    \centering
    \includegraphics[width=\textwidth]{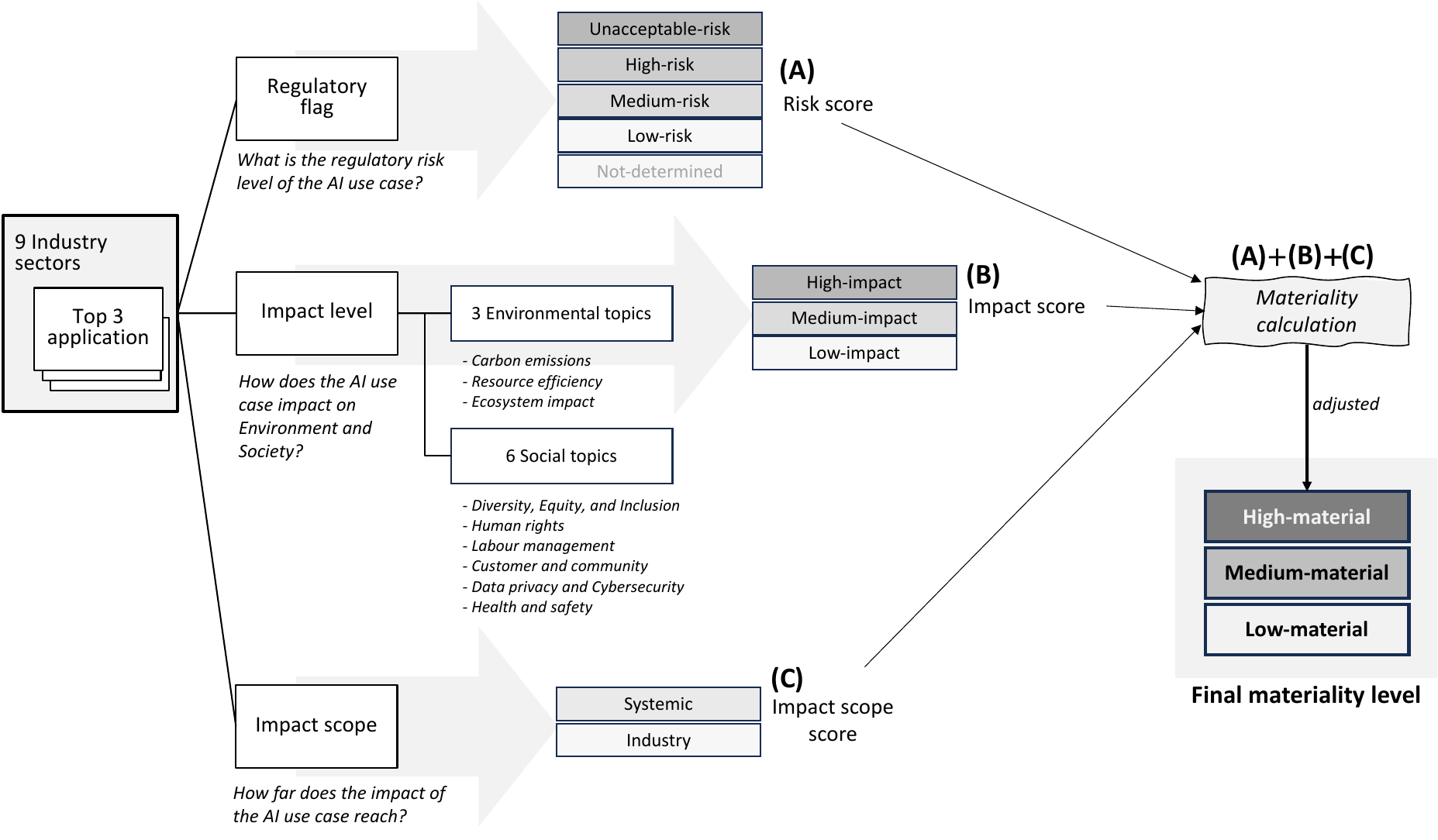}
    \caption{The AI use case analysis: three key input factors (regulatory flag, impact level and impact scope) and the final materiality level calculation process.}
    \label{fig:ai_use_case}
\end{figure*}

AI use case analysis aims to illuminate the potential benefits while identifying and mitigating inherent risks, fostering a nuanced understanding of the diverse impacts that AI can have on these industries. Figure \ref{fig:ai_use_case} shows the overall structure and analysis process of the AI use case component.

There are three input factors such as \textit{Regulatory flag}, \textit{Impact level} and \textit{Impact scope}.

\textbf{Regulatory flag} refers to the default risk of the AI use case based on the definition of AI regulations such as the EU AI Act. We have adapted the EU AI Act and defined five risk levels as follows.
\begin{itemize}
    \item \textit{Unacceptable-risk}: have significant potential to manipulate persons, vulnerabilities of specific vulnerable groups (children, disabilities), AI-based social scoring for general purposes done by public authorities.
    \item \textit{High-risk}: may create a high risk to the health and safety or fundamental rights of natural persons (e.g., biometrics, access to essential services such as credit score).
    \item \textit{Medium-risk}: fall outside the categories of unacceptable risk and high risk; instead, it is classified as an application interacting with humans.
    \item \textit{Low-risk}: exclude unacceptable/high- and medium-risk ones, demonstrate a lower level of potential harm.
    \item \textit{Not determined}: have not been definitively assessed or categorized. This could be due to various factors such as insufficient information, complexity, or ambiguity regarding the use case's impact or potential risks. 
\end{itemize}

Investors can determine the regulatory risk of the AI use case as per the definitions. However, this does not consider the specific context of the company developing or using the AI systems, and so remains at the industry level and serves as a general analysis for preliminary screening. 

\textbf{Impact level} is to assess both positive and negative impacts of the AI use case on the environment and society. 
As introduced in Section \ref{sec:background}, we used three environmental topics and six social topics as follows.

\begin{itemize}
    \item \textit{Environmental} topics: Carbon emissions, Resource efficiency, Ecosystem impact.
    \item \textit{Social} topics: Diversity/equity/inclusion, Human rights, Labour management, Customer and community, Data privacy and Cybersecurity, Health and safety.
\end{itemize}

\begin{figure*} [htb]
    \centering
    \includegraphics[width=\textwidth]{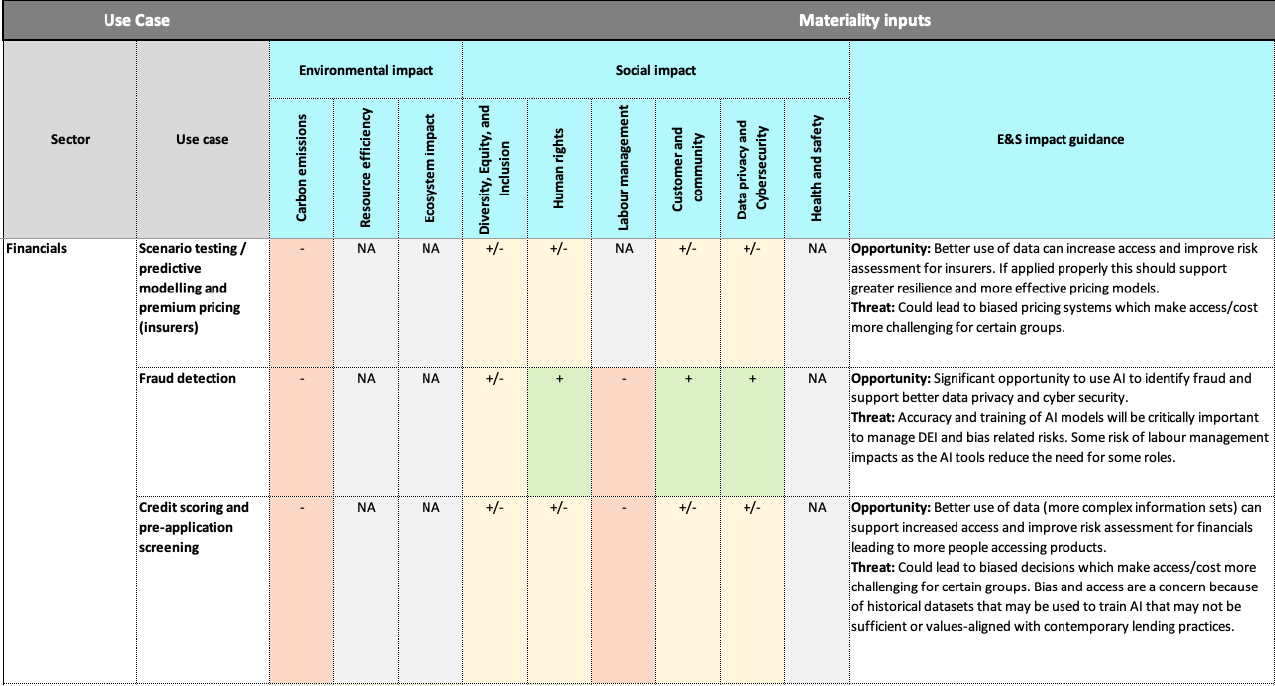}
    \caption{The screenshot of AI use case (Financial sector): positive/negative impact analysis.}
    \label{fig:ai_use_case_2}
\end{figure*}

We have defined four categories to review the impacts of the AI use case.
For example, 
investors can select (+) if the use case has positive impacts (opportunities) on the environmental and social topic; investors can select (-) if the use case has negative impacts (threats) on the environmental and social topic; investors can select (+/-) if the use case has both opportunities and threats on the environmental and social topic; investors can select (N/A) if the use case has limited impact on the environmental and social topic.
Figure \ref{fig:ai_use_case_2} shows the example analysis for financial sector AI use cases.

As both positive impact and negative impact are crucial for investors, the impact level is simply calculated based on the number of the impacted topics whether it is (+), (-), or (+/-).

We define \( N \) as the number of impacted topics and the impact level \( L \) as follows (\ref{eq:use case 1}).

\begin{equation}
L = 
\begin{cases} 
\text{"High"} & \text{if } N \ge 8 \\
\text{"Medium"} & \text{if } 3 < N \leq 7 \\
\text{"Low"} & \text{if } N \le 3 
\end{cases}
\label{eq:use case 1}
\end{equation}

\textbf{Impact scope} of AI can be categorized into several levels depending on how far the impact of AI can reach. 
Saetra \cite{saetra2021ai} proposed three levels such as Micro, Meso, and Macro.
While \textit{Micro level} is a relatively short-term and smaller impact (e.g., impact on the individual workers within a company), \textit{Macro level} refers to the broader and long term effects of AI. 

Similarly, there are three levels of impact scope used in the investment space, such as Company, Industry and Systemic.
In this study, we use investor-friendly terms and define that \textit{Company level} refers to the AI risks that have a direct impact at a company level, \textit{Industry level} is the AI risks that can have a cumulative risk to certain industries if not managed appropriately, and \textit{Systemic level} is if the AI risks are systemic and nature and pose a material risk to the economic system.
Based on the definition, we used two levels of AI impact scope (Industry and Systemic) for each AI use case. 

\textbf{Materiality score/level} is calculated using the three scores (AI risk score, AI impact score and AI impact scope).
The risk score is denoted by \( R \), the impact score by \( I \), and the impact scope score by \( S \).
The final materiality score \( F \) is calculated as follows (\ref{eq:use case 2}).

\begin{equation}
F = w_1 \cdot R + w_2 \cdot I + w_3 \cdot S
\label{eq:use case 2}
\end{equation}

where \( w_1 \), \( w_2 \), and \( w_3 \) are the respective weights for the risk score, impact score, and impact scope score.
In this study, we simply set 1 for all the weights.

The thresholds for defining the materiality level \( M \) are \( T_{\text{high}} \) and \( T_{\text{low}} \).
The initial thresholds are set as \( T_{\text{high}} = 2 \) and \( T_{\text{low}} = 1 \). However, these thresholds can be adjusted by the user to reflect different criteria.

The default materiality level \( M_{\text{default}} \) is defined as follows (\ref{eq:use case 3}).

\begin{equation}
M_{\text{default}} = 
\begin{cases} 
\text{"High"} & \text{if } F \geq T_{\text{high}} \\
\text{"Medium"} & \text{if } T_{\text{low}} \leq F < T_{\text{high}} \\
\text{"Low"} & \text{if } F < T_{\text{low}} 
\end{cases}
\label{eq:use case 3}
\end{equation}

To allow for user adjustments, the user-defined materiality level \( M_{\text{adjusted}} \) is defined as follows (\ref{eq:use case 4}).

\begin{equation}
M_{\text{adjusted}} \leftarrow M_{\text{default}}  
\label{eq:use case 4}
\end{equation}

This means that the adjusted materiality level is based on the default materiality level, automatically calculated, but the user can change the default materiality level based on their own judgment.

\subsection{ESG-AI framework: RAI Governance Indicators}

This component, \textit{RAI Governance Indicators}, has been identified from multiple sources including the company insights from the interviews, public company disclosures (e.g., annual reports), and the needs of the investors engaged with several rounds of workshop in this project.
It comprises 10 high-level indicators to assess company's overall commitment, accountability and measurement of RAI, under four different categories such as \textit{Board oversight}, \textit{RAI commitment}, \textit{RAI implementation}, and \textit{RAI metrics}.

\textbf{Board oversight} includes two indicators: Board accountability and Board capability.

\textit{Board accountability} is regarded as the central and key success criteria of traditional corporate governance \cite{moore2015neglected}.
In the context of AI governance, this indicator has been particularly chosen to strengthen company's AI management approach as many companies report AI opportunities and uptake to their boards, but this is ad hoc and lacks the consistency, according to the company engagement interviews. 
Thus indicator requires that RAI should be explicitly mentioned as part of the responsibility of the board or a relevant board subcommittee (e.g. risk committee or ESG committee) and boards should receive structured RAI reporting at least once per year but more frequently as needed. 

\textit{Board capability} refers to the ability of board to respond and act competently in the face of various problems and challenges. It is well known that good board capability leads good company performance \cite{macus2008board}. 
In the past, this topic has not received as much focus compared to studies on the implementation of AI and AI algorithms \cite{torre2019ai}.
However, board capability in AI governance plays a key role in leveraging RAI practices and has recently gained more attention from both industry and researchers. 
As evidence, 42\% of the interviewed companies had at least one director with strong capability in AI.

\textbf{RAI commitment} has three indicators as follows.

\textit{Public RAI policy} is strongly related to \textit{Transparency} principle.
Transparency and responsible disclosure are central to AI regulations, promoting user awareness of impactful AI interactions \cite{zhu2022ai, felzmann2019transparency, stahl2023embedding}.
According to our interviews with companies, 40\% had internal RAI policies, but only 10\% shared these publicly.
AI policy serves as a container for principles and guidelines that govern the development, deployment, and use of AI technologies. 
The absence or lack of visibility of an AI policy can lead to destructive consequences and/or harms to human, society and the environment due to \textit{black-boxed or ad-hoc control processes}.
Investors require companies to transparently disclosure their AI policies for right decision-making in investment. 
Accordingly, this indicator seeks to ensure that a company's AI policy should align with relevant regulations and standards (e.g. the EU AI Act, ISO/IEC 42001) and include consideration of ethics, company values, testing and transparency.

\textit{Sensitive use cases} or high-risk AI (e.g., facial recognition) should be addressed as part of the RAI policy as it can cause significant issues to health and safety or fundamental rights of natural persons.  
The EU AI Act defines comprehensive requirements for high-risk AI, including a robust risk management system, a comprehensive quality management system covering the system, model, and data aspects, meticulous record-keeping practices, and the creation of technical documents to ensure transparency. 
To comply with the legal requirements and meet AI global standards, sensitive use cases require additional oversight and approval.

\textit{RAI target} such as \% of workforce trained and reduction in RAI incidents should be clear defined and managed to support RAI policy or commitment. 
Most AI frameworks still lack the detailed guidance needed for practical implementation, particularly regarding measurable metrics related to RAI practices \cite{xia2024towards}.

\textbf{RAI implementation} relies on the following four supporting indicators to ensure responsible use of AI in daily operations. 

\textit{Dedicated RAI responsibility} means that a company need to have designated individual or function such as AI officer or similar role that has oversight for RAI.
This role is required to provide strategic guidance, ensure ethical and responsible AI practices and manage AI risks. 
An AI management committee can support a structured approach to overseeing AI initiatives by bringing together cross-functional expertise and ensuring integration across business units. 
Human agency and oversight by a dedicated role can be achieved through human-in-the-loop, human-on- the-loop, and human-in-command approaches \cite{diaz2023connecting}.
This has a link to RAI accountability which requires clear role and responsibility definitions in the accountability framework.
Lack of professional accountability mechanisms may hinder operationalising RAI in practice \cite{zhu2022ai}.

\textit{Employee awareness} is considered as a core indicator in research and in industry as well. 
ISO/IEC AI standard highlights the importance of stakeholder's awareness around AI systems. For example, it underscores that persons doing work under the organization’s control shall be aware of the AI policy, their contribution to the effectiveness of the AI management system, including the benefits of improved AI performance, and the implications of not conforming with the AI management system requirements.
Within an organization, an AI employee awareness program provides individuals with the knowledge and skills necessary to understand, develop and implement AI technologies ethically and safely.
For broader stakeholders (e.g., users), RAI awareness may include providing comprehensive information. For example, the awareness of relevant AI incidents allows users to explore AI use cases, stakeholders, and harms \cite{wang2024farsight}.

\textit{System integration} is a key indicator for assessing a company's RAI maturity. It ensures RAI practices are embedded within existing systems, avoiding isolated solutions that create silos \cite{janssen2020data, taeihagh2021governance}.
This indicator requires that RAI policy is integrated throughout existing business processes, including risk management, product development, procurement and ESG. 

\textit{AI incidents} management and report are regarded as integral part of RAI practices. 
Especially, high-risk AI (or sensitive AI use cases) should be tightly monitored and reported when there are serious incidents. 
AI incident management and reporting go beyond simply informing stakeholders. They serve as a crucial mechanism for capturing real-world failures, preventing their recurrence, and generating valuable lessons learned that benefit all stakeholders \cite{mcgregor2021preventing}.

\textbf{RAI metrics} is the last category which includes only one indicator.

\textit{RAI metrics} associated with the policy should be identified and reported externally to stakeholders.
Our company engagement highlights a clear gap in RAI measurement practices. While companies may have established RAI policies or commitments, translating them into action requires well-defined targets and a concrete execution strategy.
By entailing RAI metrics into our framework, we can incentivize companies to measure and manage their RAI performance, increase transparency by disclosing these metrics, enhance stakeholder awareness of RAI practices, and ultimately, empower investors to make more informed decisions.

Table \ref{tab:governance_indicator} presents the summary of RAI governance indicators.

\begin{table}[htb]
    \centering
    \footnotesize
    \begin{tabular}{p{0.17\textwidth}p{0.2\textwidth}p{0.55\textwidth}}
    \hline
    Category & Indicator & Description \\ 
    \hline
    Board oversight & 1. Board accountability & RAI is explicitly mentioned as part of the responsibility of the Board or a relevant Board subcommittee (e.g. risk committee or ESG committee). \\
     & 2. Board capability & At least one Director with strong technology-related experience. \\
    RAI commitment & 3. Public RAI policy & Policy should align with relevant regulations and standards (e.g., the EU AI Act, ISO/IEC 42001). \\
     & 4. Sensitive use cases & Sensitive, high-risk use cases (such as facial recognition) are addressed as part of the RAI policy. \\
     & 5. RAI target & RAI policy or commitment is supported with clear targets. \\
    RAI implementation & 6. RAI responsibility & RAI oversight can be dedicated, or part of another role or function. \\
     & 7. Employee awareness & Specific program in place to increase employee awareness of AI, alongside relevant ethical and ESG considerations.\\
      & 8. System integration & RAI policy is integrated throughout existing business processes, including risk management, product development, procurement and ESG.\\
      & 9. AI incidents & Issues and incidents related to RAI are tracked and reported internally. \\
    RAI metrics & 10. RAI metrics & RAI metrics (such as the use of AI) associated with the policy are identified and reported externally to stakeholders. \\
    \hline
    \end{tabular}
    \caption{RAI governance indicators: 4 cateogries and 10 indicators.}
    \label{tab:governance_indicator}
    \normalfont
\end{table}

RAI governance score is simply calculated using the 10 indicators as follows.

\( G_i \) represents the score for each of the 10 governance indicators and \( F \) denotes the final governance score.
Currently, each of the indicators is assigned an equal weight of 1. Thus, the formula with weights can be expressed as follows (\ref{eq:governance indicator 1}).

\begin{equation}
F = \sum_{i=1}^{10} w_g G_i    
\label{eq:governance indicator 1}
\end{equation}

where \( w_g = 1 \) is the weight for each indicator.

For the final RAI governance level, the governance level \( L \) is determined as follows (\ref{eq:governance indicator 2}).

\begin{equation}
L = 
\begin{cases} 
\text{"High"} & \text{if } F \ge 8 \\
\text{"Medium"} & \text{if } 3 < F \leq 7 \\
\text{"Low"} & \text{if } F \le 3 
\end{cases}    
\label{eq:governance indicator 2}
\end{equation}

\subsection{ESG-AI framework: RAI Deep Dive Assessment}

\textit{RAI Deep Dive} is to facilitate detailed analysis and engagement with company management on AI governance and RAI practices. 
AI ethics principles encompass key values and guidelines that address RAI development and deployment. 
Investors can use this assessment for a systematic evaluation of fairness, transparency, accountability, privacy, and more, contributing to a holistic understanding of how well a company adheres to ethical standards in its AI practices.
This assessment has been developed based on the RAI question bank \cite{lee2023qb4aira} and metric catalogue \cite{xia2024towards}, which draws on insights and standards from key regulatory bodies, standard organisations and stakeholder groups, such as the EU AI Act, NIST AI Risk Management Framework, ISO AI Standard (ISO/IEC 42001) and other industry AI risk frameworks.

\begin{figure*} [htb]
    \centering
    \includegraphics[width=\textwidth]{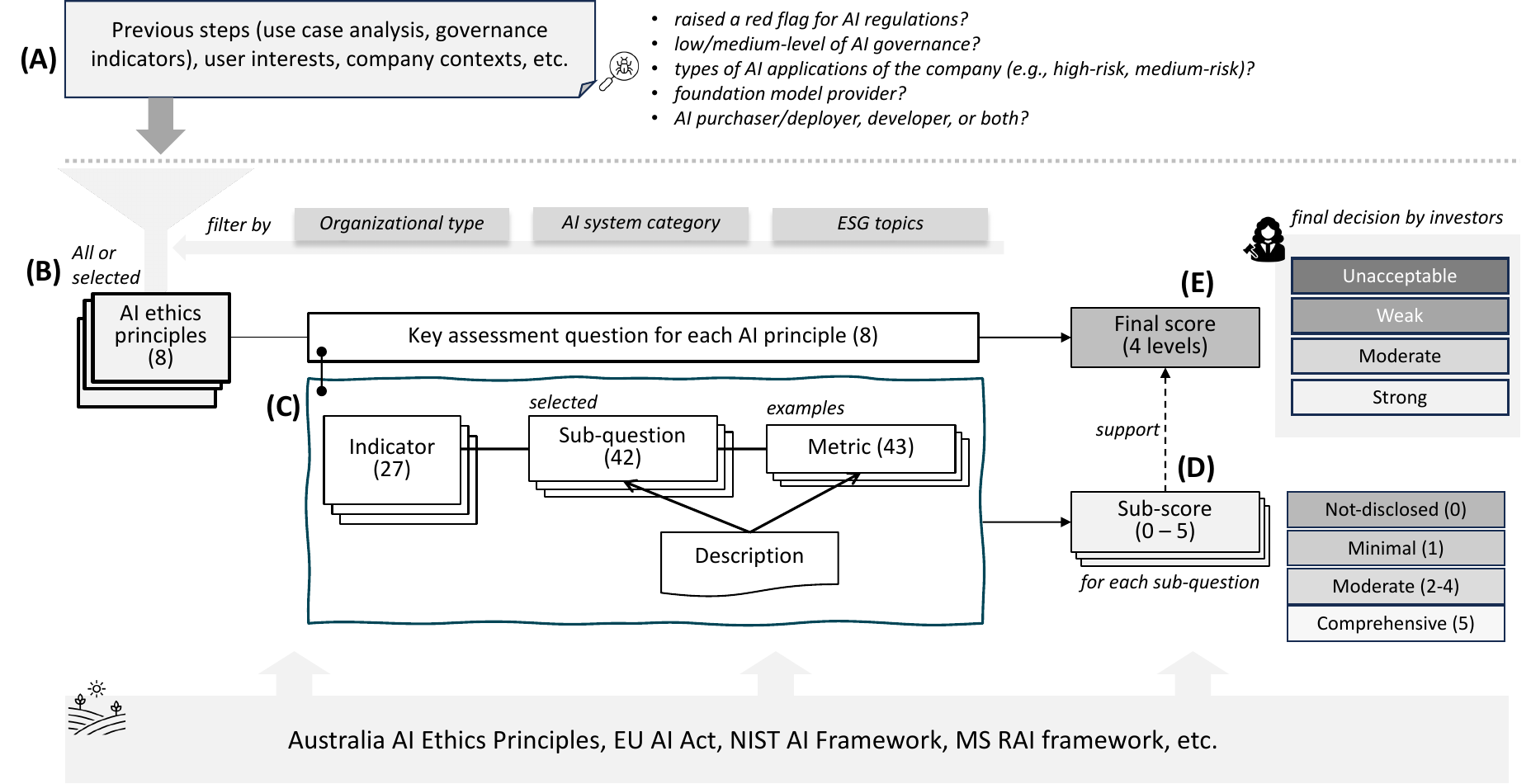}
    \caption{The Deep Dive Assessment: key elements and the overall process.}
    \label{fig:deep_dive}
\end{figure*}

\textbf{Deep dive assessment process.} Figure \ref{fig:deep_dive} shows the structure and describes how user can use it. It enables an in-depth assessment of RAI pracitces, but investors can flexibly undertake research on specific ESG concerns or use cases and tailor the questions based on your ESG interests or by material principles. 

There are 42 assessment questions distributed among the 8 AI ethics principles with 27 specific indicators. Table \ref{tab:deep dive_accountability} shows example questions, indicators, and metrics for Accountability principle.

\begin{table}[htb]
    \centering
    \footnotesize
    \begin{tabular}{p{0.2\textwidth}p{0.15\textwidth}p{0.35\textwidth}p{0.2\textwidth}}
    \hline
    Key question & Indicator & Sub-question & Metric \\
    \hline
    Does the company have designated responsibility for AI and RAI within the organisation? & Risk management & Does the company establish methods and metrics to quantify and measure the risks associated with its AI systems? & Number of AI risk metrics (e.g., risk exposure index, risk severity score) \\
     & AI incident management & Does the company have a clear reporting system or process in place for serious AI incidents to inform external stakeholders (e.g., market surveillance authorities, communities) beyond the company? & Number of AI incidents informed to external stakeholders \\
      & Accountability framework & Does the company have an accountability framework to ensure that AI related roles and responsibilities are clearly defined? & Percentage of defined AI roles and responsibilities \\
    \hline
    \end{tabular}
    \caption{Example questions and metrics (Accountability).}
    \label{tab:deep dive_accountability}
    \normalfont
\end{table}

\textbf{Selection of assessment questions.} Users can consider potential areas for further review identified in the previous steps, including concerns related to AI regulations, high-risk applications, and specific areas with lower AI governance scores (Figure \ref{fig:deep_dive}- (A)). Our framework provides three options for filtering such as \textit{Organizational type}, \textit{AI system category} and \textit{ESG topics}.

\textit{Organizational type} is categorized by different parties involved in the operation of AI systems. Lee et al. \cite{lee2021deep} defined three parties including the creator of the AI, the user of the AI (who is the purchaser of the algorithm), and the targets of the AI. 
As our framework is to assess companies, we have selected the first and second parties, namely, \textit{AI developer} and \textit{AI purchaser} respectively. We also added \textit{both, AI developer/purchaser} as another option.

\textit{AI system category}, such as the one proposed in the EU AI Act, categorizes AI systems into different risk levels like high-risk and low-risk, along with foundation model. This can help users select targeted questions to assess the risk profiles of companies using AI.

\begin{figure*} [htb]
    \centering
    \includegraphics[width=\textwidth]{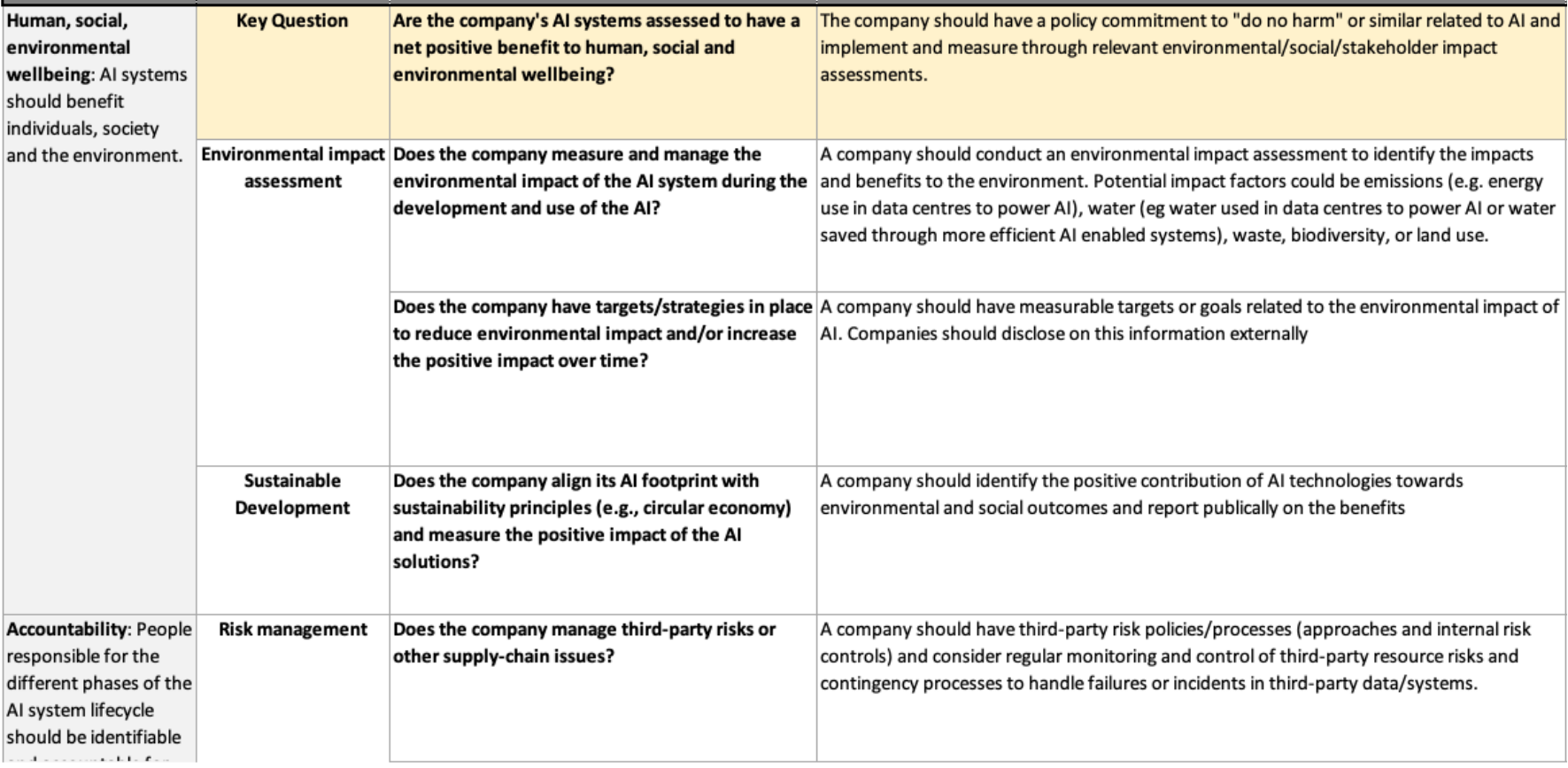}
    \caption{The screenshot of RAI deep dive assessment: four sub-questions are selected after applying ESG filter ("Carbon emissions").}
    \label{fig:ESG filter}
\end{figure*}

\textit{ESG topics} encompass a set of 12 standardized aspects that are chosen by participating investors. This enables users to conduct a deep dive assessment of associated RAI practices of the company at the ESG topic level.
For example, \textit{Carbon emissions}, the first environmental topic and four sub-questions have a connection as shown in Figure \ref{fig:ESG filter}.
The first three questions are directly related to the environmental impacts of AI, which fall under \textit{Human, societal, and environmental wellbeing} principle. The last question broadly includes third-party risks including all ESG aspects.

This structured approach ensures that all critical aspects of ESG are systematically reviewed, allowing for a thorough analysis of how a company integrates responsible AI practices into its broader ESG strategy. 

As demonstrated, users can adopt the 8 leading principle questions (all or some selected principles by filtering) in company engagement and RAI analysis (Figure \ref{fig:deep_dive}- (B)) and conduct a deep dive at the principle level.

\textbf{Role of sub-questions and scoring method.}
Utilizing the sub-questions and completing a full assessment help investors 
gain a deeper understanding of the state of RAI practices (Figure \ref{fig:deep_dive}- (C)). 
This assessment mainly evaluates companies' exposure to high-risk AI use cases and weaknesses in their RAI governance through a series of sub-questions. Each sub-question awards a sub-score to contribute to a final score (Figure \ref{fig:deep_dive}- (D)).
We used a Likert scoring system which is a popular method used in both academia and industry. Our 6-point scale is based on a general 5-point scale (score 1 - 5). We have also included \textit{0 point} to provide more varieties of options \cite{joshi2015likert}.

Score 0 refers to \textit{Not-disclosed} which means the company does not provide any evidence against the assessment question to prove their RAI practice implementation.
Score 1 is awarded as the company provides \textit{minimal} information that is  insufficient for investors to understand or lacks transparency.
Score 2-4 is given if the company shows a \textit{moderate} level of disclosure, which is reasonably detailed and provides a fair understanding for investors.
Score 5 means that the company provides \textit{comprehensive} and exemplary disclosure, ensuring that investors no longer require extra information or evidence. The scoring function is as follows (\ref{eq:deep dive 1}).
 
\begin{equation}
\text{Score} = 
\begin{cases} 
0 & \text{if no evidence is provided (Not-disclosed)} \\
1 & \text{if minimal information is provided (Minimal)} \\
2-4 & \text{if moderate level of disclosure is provided (Moderate)} \\
5 & \text{if comprehensive and exemplary disclosure is provided (Comprehensive)}
\end{cases}
\label{eq:deep dive 1}
\end{equation}

\textbf{Guide metrics and its usage.}
To address the issue of limited visibility of company's RAI policies and metrics highlighted in previous sections, 43 guide metrics are included as a key area for which investors strongly encourage companies to increase public disclosure.
These metrics are collected from the company engagement, literature review, the participating investors and AI regulations and frameworks (e.g., the EU AI Act) and distributed across the sub-questions.
This study does not assign specific numerical values to sub-question scoring.
However, it provides a reference point to facilitate scoring, aiding investors in assessing the relative performance of a company in the context of responsible AI practices. 
We particularly suggest 6 mandatory metrics to comply with AI regulations (e.g., the EU AI Act) for High-risk AI and foundation models as follows (Table \ref{tab:metics}).

\begin{table}[htb]
    \centering
    \footnotesize
    \begin{tabular}{p{0.25\textwidth}p{0.42\textwidth}p{0.25\textwidth}}
    \hline
    Metric & Description/measurement guide & Target \\
    \hline
    Energy usage & Include energy consumption of data centres, AI models, AI systems, AI training pipelines, AI devices, etc. \textit{(X = A, A = Amount of energe used for AI, X >= 0 The smaller, the better)} & high-risk AI, foundation model \\
    Greenhouse gas emission & Include gas emission of data centres, AI models, AI systems, AI training pipelines, AI devices, etc. \textit{(X = A, A = Amount of gas emission for AI, X >= 0 The smaller, the better)} & high-risk AI, foundation model \\
    Tonnes of waste generated/saved & Include amount of waste generated or saved during development and operation, from data centres, AI systems, AI devices, etc. \textit{(X = A, A = Amount of waste generated/saved for AI, X >= 0 The smaller/bigger, the better)} & high-risk AI, foundation model \\
    AI system performance & Evaluate whether the information processed and analyzed by AI systems is free from errors, inconsistencies, or biases. \textit{(it can include metrics such as accuracy ((true positive + true negative) / total predictions), precision (true positive / (true positive + false positive)), recall (true positive / (true positive + false negative)), fscore (2 x ((precision x recall) / (precision + recall)))} & high-risk AI, foundation model \\
    Size of AI system (model) & Consider the cost for model training, including AI model/data size. & foundation model \\
    Time to AI model training & Measure the time to train AI models. \textit{(X = A, A = time spent on model training, X >= 0, The smaller, the better)} & foundation model \\
    \hline
    \end{tabular}
    \caption{Mandatory metrics suggested by this study: this provides alignment with AI regulations such as the EU AI Act.}
    \label{tab:metics}
    \normalfont
\end{table}

\textbf{Key questions and final decision.}
The lead questions, or key questions, are designed to assess the company at the principle level and provide consolidated decisions that accommodate all requirements from the sub-questions.
These questions serve as an overarching question to interpret the core concepts of the AI principle, ensuring high-level insights on the company's RAI practices.
Investors can decide the final score/level of the company reflecting the score from the sub-questions. 

The final decision represents one of four levels such as \textit{Unacceptable}, \textit{Weak}, \textit{Moderate}, and \textit{Strong} (Figure \ref{fig:deep_dive}- (E)).
In this study, our framework empowers investors to determine the final score based on their own judgment. 
The following provides simple scoring guidance using average score of the sub-questions, to support their decision-making.

The average score of the sub-questions is calculated as follows (\ref{eq:deep dive 2}).

\begin{equation}
\text{Average Score} = \frac{\sum_{i=1}^{n} \text{SubQuestionScore}_i}{n}
\label{eq:deep dive 2}
\end{equation}

where \( n \) is the number of sub-questions, and \(\text{SubQuestionScore}_i\) is the score of the \(i\)-th sub-question.

The final level of the lead question is determined based on the average score (\ref{eq:deep dive 3}).

\begin{equation}
\text{Final Level} = 
\begin{cases} 
\text{Strong} & \text{if Average Score } \geq 4.5 \\
\text{Moderate} & \text{if } 3 \leq \text{Average Score} < 4.5 \\
\text{Weak} & \text{if } 1.5 \leq \text{Average Score} < 3 \\
\text{Unacceptable} & \text{if Average Score} < 1.5 
\end{cases}    
\label{eq:deep dive 3}
\end{equation}

\section{Discussion} \label{sec:discussion}

\subsection{User interest and feedback} 

Throughout the project, we conducted several workshops involving investors (potential users) and senior industrial experts, fostering \textit{iterative testing and enhancements}. 
These engaged participants actively contributed to the continuous refinement of the framework, offering valuable insights into its usability, particularly focusing on the clarity and practical applicability of the framework. 
The feedback underscored the importance of framework comprehensibility, real-world utility, and the effectiveness of the guidance provided for each component. 
This iterative approach ensured that the risk assessment questions within the framework were meticulously tailored to meet the diverse needs of its users.
Consequently, the framework was widely acknowledged as a highly effective and pragmatic tool for assessing AI and ESG risks among investors.

In April 2024, we released the framework along with the final project report. Within the first 24 hours, it received coverage from 23 media outlets, and the number is still increasing.
Moreover, within a week of the release, more than 1,000 people downloaded our final report from the website, and around 100 downloaded the framework toolkit, despite the gated website requiring personal information.
Our framework has also received positive feedback from the audience. 
We had several follow-up meetings with various investment companies and garnered significant attention from them.

In early May, we promoted the framework at one of the world largest investor conferences. Many attendees, including investors, researchers, and industry practitioners, visited our booth and expressed their interest.

We also gained significant attention from social media users since we posted our framework. Here are some comments from them:

\textit{"Great to see this media coverage for your publication on Responsible AI and ESG integration. Such an important topic area. This work will provide valuable insights into addressing key challenges faced by executives and boards."}

\textit{"...it sounds well-served in going beyond a principles-only approach to implementing RAI."}


\subsection{Practical implications}

\textbf{Analysis on AI use cases in different sectors.}
We have analyzed 27 AI use cases across 9 sectors, assessing regulatory risk (unacceptable/high/medium/low/not-determined), environmental and social impacts (9 topics), and impact scope (industry/systemic) with input from two participating investors.

\begin{figure*} [htb]
    \centering
    \includegraphics[width=0.9\textwidth]{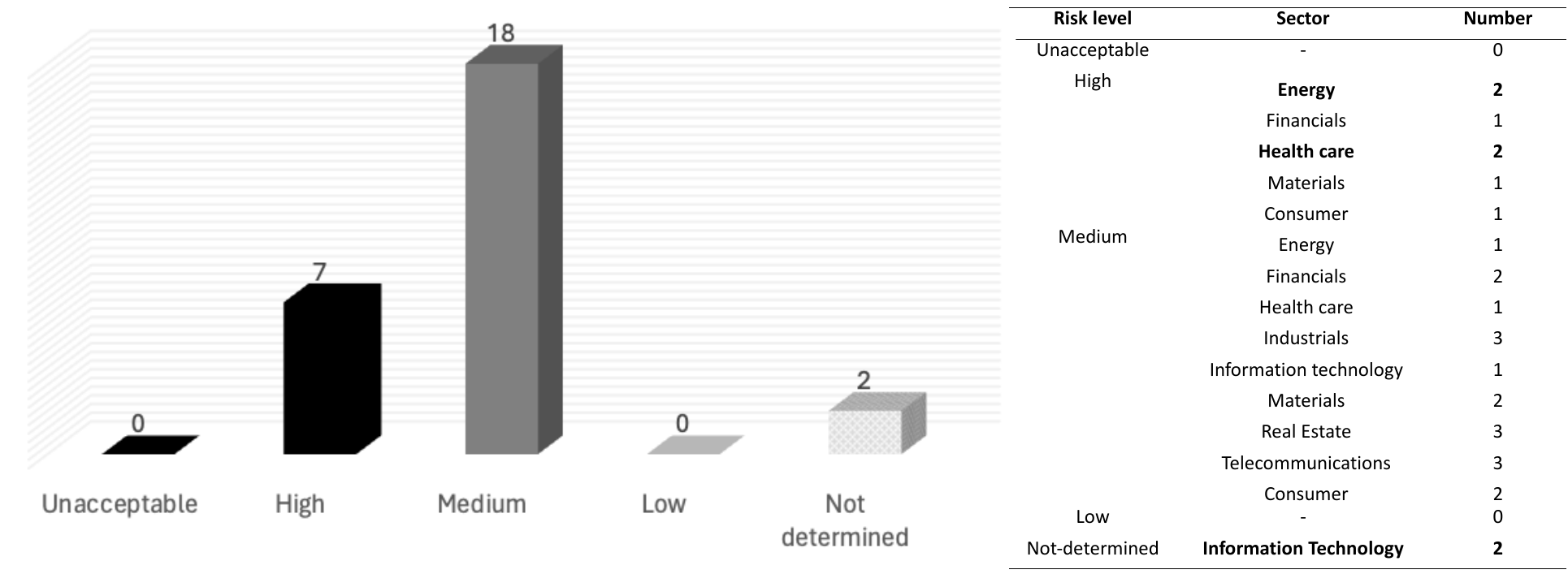}
    \caption{The regulatory risk level of the AI use cases.}
    \label{fig:risk level}
\end{figure*}

Figure \ref{fig:risk level} shows that most AI use cases are considered medium-risk, with no unacceptable or low-risk use cases identified. However, both the Energy and Healthcare sectors include two high-risk AI use cases. In the Energy sector, \textit{Predictive infrastructure maintenance} and \textit{Grid management and energy optimisation} are high-risk as they pertain to \textit{Critical infrastructure}. Similarly, in the Healthcare sector, \textit{Health research/testing} and \textit{Clinical care} use cases are also categorized as \textit{Critical infrastructure} and involve biometrics, a high-risk area defined by regulations. These use cases must adhere to regulatory requirements (e.g., the EU AI Act), which include a robust risk management system, a comprehensive quality management system covering the system, model, and data aspects, meticulous record-keeping practices, and the creation of technical documents to ensure transparency.
Two use cases, \textit{Product development} and \textit{Automation} in the Information Technology sector, are classified as \textit{not-determined} due to the need for further information to assess the risk level, given the implementation of AI systems across different processes and functions.

\begin{figure*} [htb]
    \centering
    \includegraphics[width=0.9\textwidth]{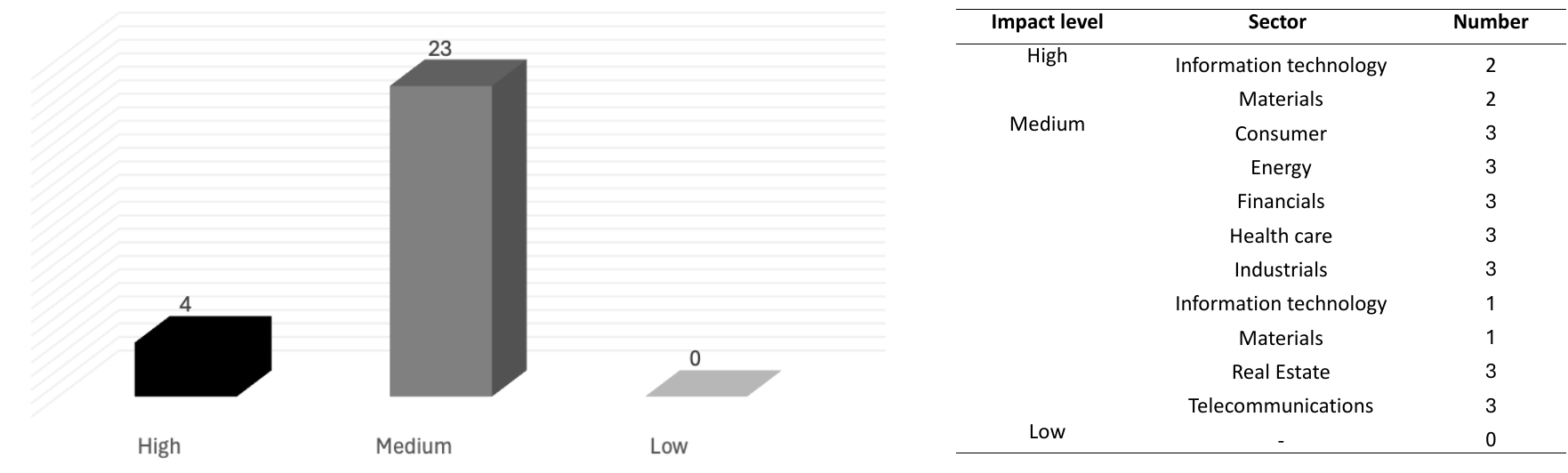}
    \caption{The environmental and social impact of the AI use cases.}
    \label{fig:impact level}
\end{figure*}

Figure \ref{fig:impact level} illustrates how AI use cases influence the Environment and Society. Our analysis encompassed 9 topics, excluding governance topics, as these have been evaluated at the industry level. As shown in the table in the figure, four AI use cases have a high impact on both the Environment and Society. Particularly, use cases in the Information sector impact all topics, while those in the Materials sector broadly impact the Environment and Society (except for the "Diversity, Equity, and Inclusion" topic). There are no low-impact AI use cases, indicating that all analyzed use cases affect more than three topics. 

\begin{figure*} [htb]
    \centering
    \includegraphics[width=0.9\textwidth]{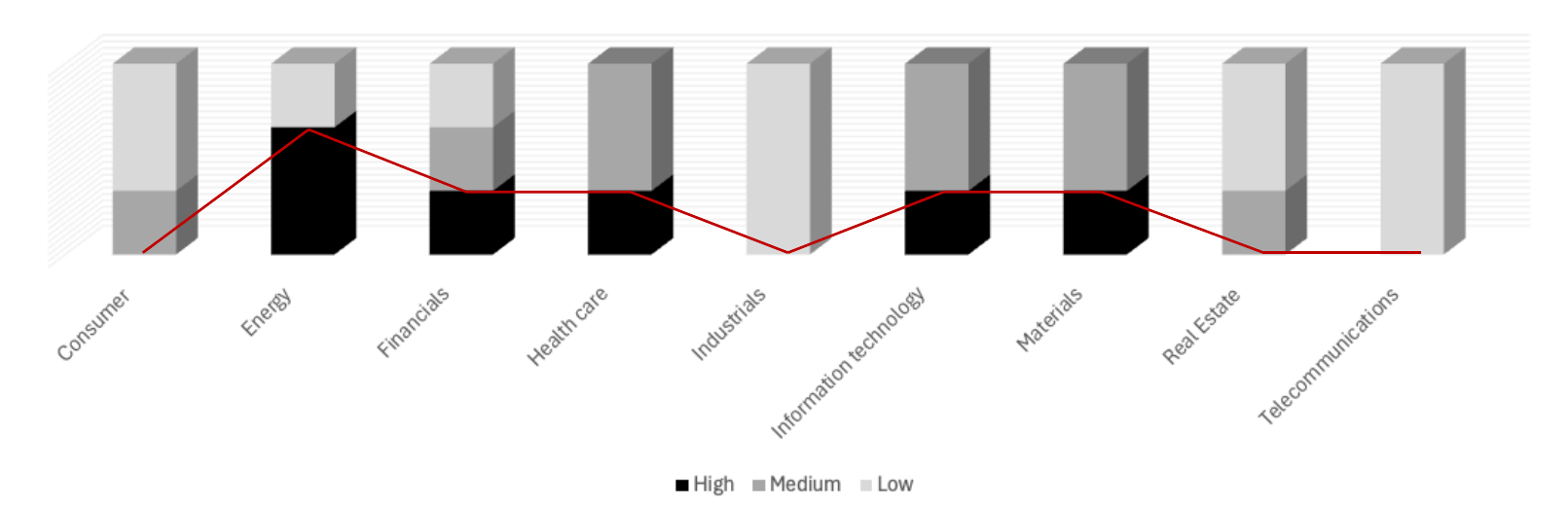}
    \caption{The materialist level of the AI Use Cases: while Energy sector AI applications are the highest material, Telecommunications AI applications are the lowest meatrial.}
    \label{fig:materiality level}
\end{figure*}

The impact scope analysis shows that 8 AI use cases in 6 sectors have a broad, systemic impact. For example, in the financial sector, \textit{Credit scoring and pre-application screening} has systemic risks that pose material threats to the economic system. Similarly, \textit{Clinical care} in the healthcare sector potentially has destructive threats affecting entire industries and society. These systemic risks may have direct and indirect impacts with longer-term time horizons, necessitating robust management support for companies using/adopting these AI applications.

The final materiality level of the use cases can be determined based on the aforementioned risk and impact analysis.
Figure \ref{fig:materiality level} provides an overview of the materiality levels across 9 sectors. As shown, five sectors include high-material use cases, with the Energy sector having two high-level cases. Conversely, the Telecommunications sector is relatively low in materiality from an investor perspective. This result indicates which industry sectors and AI applications should be more closely reviewed and considered when making investment decisions. From the company's perspective, they need to carefully adopt and develop AI systems with ongoing monitoring, reporting, and control to minimize negative consequences and improve ESG performance \cite{saetra2023ai}.

\begin{figure*} [htb]
    \centering
    \includegraphics[width=0.75\textwidth]{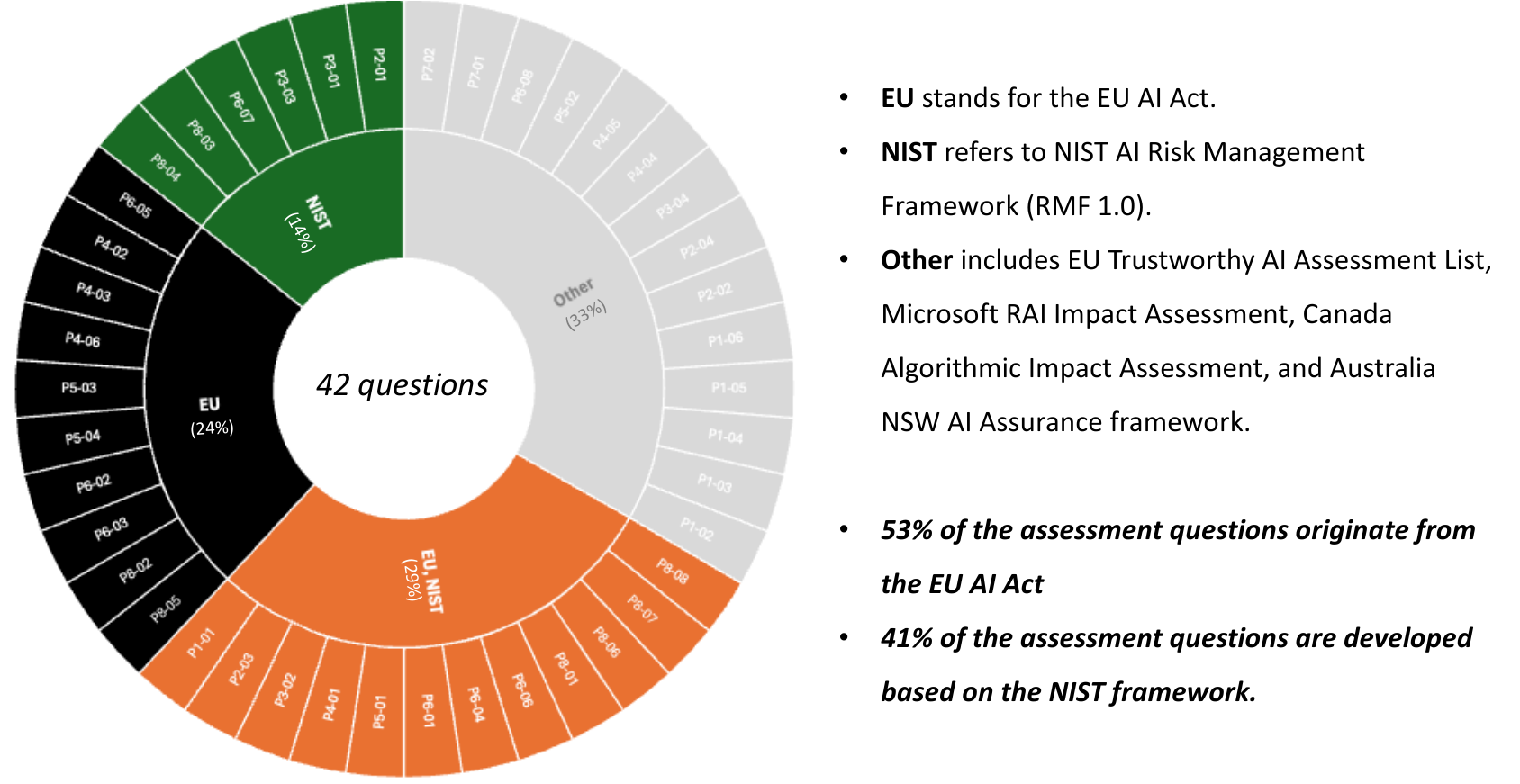}
    \caption{42 questions in Deep Dive Assessment and the alignment with the EU AI Act, NIST framework and others.}
    \label{fig:framework alignment}
\end{figure*}

\textbf{Alignment with global and industry requirements.}
As described earlier, our framework is designed with a robust foundation, drawing insights and standards from key regulatory bodies such as the EU AI Act, the NIST AI Risk Management Framework (RMF), and other industry AI frameworks (Figure \ref{fig:framework alignment}). 
For example, the deep dive assessment comprises 67\% of assessment questions developed based on the EU AI Act and NIST framework; this includes 14\% of NIST only questions, 24\% of EU only questions, and 29\% of both EU and NIST questions.

The deep dive assessment is particularly developed encompassing legal requirements for critical AI systems such as high-risk AIs and foundation models.
For high-risk AI, users can select 17 mandatory questions addressing legal requirements, complemented by 5 optional questions that delve into broader AI ethics principles.
When assessing foundation model providers, users can choose from 13 mandatory questions and consider 8 additional questions.
This enables through the filters provided by our framework tool.

By incorporating requirements and recommendations from these authoritative sources, our assessment is aligned with internationally recognized regulations and standards, ensuring a thorough evaluation of AI systems across legal, ethical, and risk management dimensions. 
This approach not only helps investors understand regulatory compliance but also ensures a holistic and well-rounded assessment that addresses the multifaceted challenges associated with AI deployments.

\begin{table}[htb]
    \centering
    \footnotesize
    \begin{tabular}{p{0.07\textwidth}p{0.05\textwidth}p{0.05\textwidth}p{0.05\textwidth}p{0.05\textwidth}p{0.05\textwidth}p{0.05\textwidth}p{0.05\textwidth}p{0.05\textwidth}p{0.05\textwidth}p{0.05\textwidth}p{0.05\textwidth}p{0.05\textwidth}}
    \hline
    Principle & \multicolumn{3}{c}{Environment} & \multicolumn{6}{c}{Social} & \multicolumn{3}{c}{Governance} \\
     & E1 & E2 & E2 & S1 & S2 & S3 & S4 & S5 & S6 & G1 & G2 & G3 \\
    \hline
    HSE & \cellcolor{gray!30}3 & \cellcolor{gray!30}3 & \cellcolor{gray!30}3 & \cellcolor{gray!40}4 & \cellcolor{gray!40}4 & \cellcolor{gray!30}4 & \cellcolor{gray!30}3 & \cellcolor{gray!30}3 & \cellcolor{gray!30}3 & \cellcolor{gray!10}1 & \cellcolor{gray!10}1 & \cellcolor{gray!20}2 \\
    HV & 0 & 0 & 0 & \cellcolor{gray!40}4 & \cellcolor{gray!40}4 & \cellcolor{gray!40}4 & \cellcolor{gray!40}4 & \cellcolor{gray!40}4 & \cellcolor{gray!40}4 & \cellcolor{gray!10}1 & \cellcolor{gray!20}2 & \cellcolor{gray!10}1 \\
    FAR & 0 & 0 & 0 & \cellcolor{gray!40}4 & \cellcolor{gray!40}3 & \cellcolor{gray!20}2 & \cellcolor{gray!10}1 & 0 & 0 & 0 & 0 & \cellcolor{gray!10}1 \\
    PRV & 0 & 0 & 0 & 0 & \cellcolor{gray!40}4 & \cellcolor{gray!20}2 & \cellcolor{gray!40}4 & \cellcolor{gray!60}6 & 0 & 0 & \cellcolor{gray!40}4 & 0 \\
    REL & 0 & \cellcolor{gray!10}1 & \cellcolor{gray!10}1 & \cellcolor{gray!10}1 & \cellcolor{gray!10}1 & 0 & \cellcolor{gray!20}2 & \cellcolor{gray!40}4 & \cellcolor{gray!10}1 & \cellcolor{gray!20}2 & \cellcolor{gray!10}1 & 0 \\
    TRN & 0 & 0 & 0 & 0 & 0 & \cellcolor{gray!60}6 & \cellcolor{gray!60}6 & 0 & 0 & 0 & \cellcolor{gray!10}1 & \cellcolor{gray!60}6 \\
    CON & 0 & 0 & \cellcolor{gray!10}1 & \cellcolor{gray!10}1 & \cellcolor{gray!20}2 & \cellcolor{gray!10}1 & \cellcolor{gray!10}1 & 0 & \cellcolor{gray!10}1 & 0 & 0 & 0 \\
    ACC & \cellcolor{gray!10}1 & \cellcolor{gray!10}1 & \cellcolor{gray!10}1 & \cellcolor{gray!10}1 & \cellcolor{gray!10}1 & \cellcolor{gray!20}2 & \cellcolor{gray!10}1 & \cellcolor{gray!10}1 & \cellcolor{gray!10}1 & \cellcolor{gray!40}4 & \cellcolor{gray!50}5 & \cellcolor{gray!20}2 \\
    \hline   
    \multicolumn{13}{l}{AI ethics principles:} \\
    \multicolumn{13}{l}{HSE: human, societal, environmental wellbeing / HV: human-centred value / FAR: Fairness / PRV: privacy and security} \\
    \multicolumn{13}{l}{REL: reliability and safety / TRN: transparency and explainability / CON: contestability / ACC: accountability} \\ \\

    \multicolumn{13}{l}{ESG topics:} \\
    \multicolumn{13}{l}{E1: carbon emissions / E2: resource efficiency / E3: ecosystem impact} \\
    \multicolumn{13}{l}{S1: diversity, equity, and inclusion / S2: human rights / S3: labour management / S4: customer and community} \\
    \multicolumn{13}{l}{S5: data privacy and cybersecurity / S6: health and safety} \\
    \multicolumn{13}{l}{G1: board and management / G2: policy / G3: disclosure and reporting} \\
    \hline 
    \end{tabular}
    \caption{Mapping between RAI principles and ESG topics.}
    \label{tab:ESG mapping}
    \normalfont
\end{table}

\textbf{Alignment with ESG.}
Our framework involves mapping the assessment questions to ESG considerations, allowing for a comprehensive evaluation of how a company's RAI practices align with broader sustainability goals and ethical principles.

Table \ref{tab:ESG mapping} presents the relationship between RAI principles and ESG topics, which is generated by analysing the assessment questions and the link to the ESG topics.

This table clearly shows that \textit{Transparency} is strongly related to \textit{Social aspects (labour and customer management)} and \textit{Disclosure and reporting in Governance.}
Likewise, \textit{Privacy and security} principle is tightly coupled with \textit{Data privacy and cybersecurity} which shows there are large overlaps between them.
\textit{Accountability} principle, however, is mainly related to \textit{Governance topics in ESG} but rarely mapped with other ESG topics.
\textit{Human-centred values} and \textit{Fairness} principles have connections with \textit{Social topics} as these principles focus on the similar areas such as diversity and inclusion and human rights.







\section{Conclusion} \label{sec:conclusion}

In this paper, we have developed and presented a comprehensive ESG-AI framework designed to guide investors in integrating ESG considerations with AI practices. Our framework addresses the critical need for RAI implementation by providing practical tools and guidelines that go beyond conceptual discussions, thus filling a significant gap in the current landscape.

Through our detailed analysis and comparison with existing frameworks, we have highlighted the unique value of our approach in comprehensively covering both ESG topics and AI ethics principles. By leveraging Australia's eight AI ethics principles and incorporating twelve key ESG topics, our framework offers a robust and actionable toolkit for investors.
We believe that the adoption of our ESG-AI framework will enable investors to make informed, ethical, and sustainable decisions, ultimately contributing to a more responsible and future-proof AI ecosystem. 

Future work can expand on this foundation by refining and testing the framework in various industry contexts, ensuring its adaptability and effectiveness across different sectors.
As a first step, we will gather feedback from individuals who have downloaded our framework and toolkit by conducting a user survey. This will provide valuable insights from potential users, including both investors and companies, about real-world applications.

\bibliographystyle{unsrt}  
\bibliography{main}

\end{document}